\documentclass[lettersize,journal]{IEEEtran}
\usepackage{amsmath,amsfonts}
\usepackage{algorithmic}
\usepackage{algorithm}
\usepackage{array}
\usepackage[caption=false,font=normalsize,labelfont=sf,textfont=sf]{subfig}
\usepackage{textcomp}
\usepackage{stfloats}
\usepackage{url}
\usepackage{verbatim}
\usepackage{graphicx}
\usepackage{cite}
\usepackage{threeparttable}
\usepackage{hyperref}
\hypersetup{
    colorlinks=true,
    linkcolor=blue,
    citecolor=blue,
    urlcolor=blue
}
\usepackage{booktabs}
\usepackage[table]{xcolor}
\usepackage{multirow}
\usepackage{graphicx}
\usepackage{caption}
\usepackage{lipsum}
\begin{document}

\title{All-in-One Video Restoration under Smoothly Evolving Unknown Weather Degradations}

\author{Wenrui Li,~\IEEEmembership{~Member,~IEEE}
        Hongtao Chen,  
        Yao Xiao,
        Wangmeng Zuo,~\IEEEmembership{~Senior Member,~IEEE}\\
        Jiantao Zhou,~\IEEEmembership{~Senior Member,~IEEE}
        Yonghong Tian,~\IEEEmembership{~Fellow,~IEEE}
        Xiaopeng Fan,~\IEEEmembership{~Senior Member,~IEEE}
\thanks{This work was supported in part by the National Key R\&D Program of China (2023YFA1008501) and the National Natural Science Foundation of China (NSFC) under grant 624B2049 and U22B2035. (Corresponding author: Xiaopeng Fan.)}
\thanks{Wenrui Li, Hongtao Chen, Yao Xiao, Wangmeng Zuo and Xiaopeng Fan are with the Department of Computer Science and Technology, Harbin Institute of Technology, Harbin 150001, China. (e-mail: liwr618@163.com; ht166chen@163.com; 25S003001@stu.hit.edu.cn; wmzuo@hit.edu.cn; fxp@hit.edu.cn).}
\thanks{Jiantao Zhou is with the State Key Laboratory of Internet of Things for Smart City and the Department of Computer and Information Science, University of Macau, Macau, China (e-mail: jtzhou@um.edu.mo).}
\thanks{Yonghong Tian is with the School of AI for Science, the Shenzhen Graduate School, Peking University, Shenzhen, China, the Peng Cheng Laboratory, Shenzhen, China, and also with the School of Computer Science, Peking University, Beijing, China (e-mail: yhtian@pku.edu.cn).}
}

\markboth{Journal of \LaTeX\ Class Files,~Vol.~14, No.~8, August~2021}%
{Shell \MakeLowercase{\textit{et al.}}: A Sample Article Using IEEEtran.cls for IEEE Journals}


\maketitle

\begin{abstract}

All-in-one image restoration aims to recover clean images from diverse unknown degradations using a single model. But extending this task to videos faces unique challenges.  Existing approaches primarily focus on frame-wise degradation variation, overlooking the temporal continuity that naturally exists in real-world degradation processes. 
In real world, degradation types and intensities evolve smoothly over time, and multiple degradations may coexist or transition gradually. 
In this paper, we introduce the Smoothly Evolving Unknown Degradations (SEUD) scenario to simulate such real-world processes, where both the active degradation set and degradation intensity change continuously over time. 
To support this scenario, we design a flexible synthesis pipeline that generates temporally coherent videos with single, compound, and evolving degradations.
To address the challenges in the SEUD scenario, we propose an all-in-One Recurrent Conditional and Adaptive prompting Network (ORCANet). First, a Coarse Intensity Estimation Dehazing (CIED) module estimates haze intensity using physical priors and provides coarse dehazed features as initialization. Second, a Flow Prompt Generation (FPG) module extracts degradation features. 
FPG generates both static prompts that capture segment-level degradation types and dynamic prompts that adapt to frame-level intensity variations.
Furthermore, a label-aware supervision mechanism improves the discriminability of static prompt representations under different degradations. Extensive experiments show that ORCANet achieves superior restoration quality, temporal consistency, and robustness over image and video-based baselines. Code is available at https://github.com/Friskknight/ORCANet-SEUD.

\end{abstract}


\section{Introduction}
\IEEEPARstart{A}{ll-in-one} image restoration (AiOIR) aims to recover clean images with a single model under diverse and unknown degradations \cite{aiowea2020Li, transweather2022Jose, air2022Li, PIPli2023, wdiffusion2023ozan, PerIR25zhang, UniUIR25zhang, relation25Li}. Unlike task-specific methods\cite{MFD25xiao, Ye2022ECCVDensity, Song2022VITDehaze, D42022yang, BMVC2024Kirillova, UCL24wang, sid2021yang, ad2016you, hearain2019li, desnow2018liu, Qian2018Raindrop, Quan2021OneGo, Chen2020JSTASR, deepse2021zhang, dis2023quan, famamba2024xiao, gou2022multi, TTST24xiao, gou2023rethinking, AFD23zhang}, which require separate models for denoising, deblurring, or dehazing, all-in-one approaches handle multiple degradation types in a single framework. This generality is important for real-world scenarios, where degradations are often diverse, composite, and unpredictable.


\begin{figure}
    \centering
    \includegraphics[width=1\linewidth]{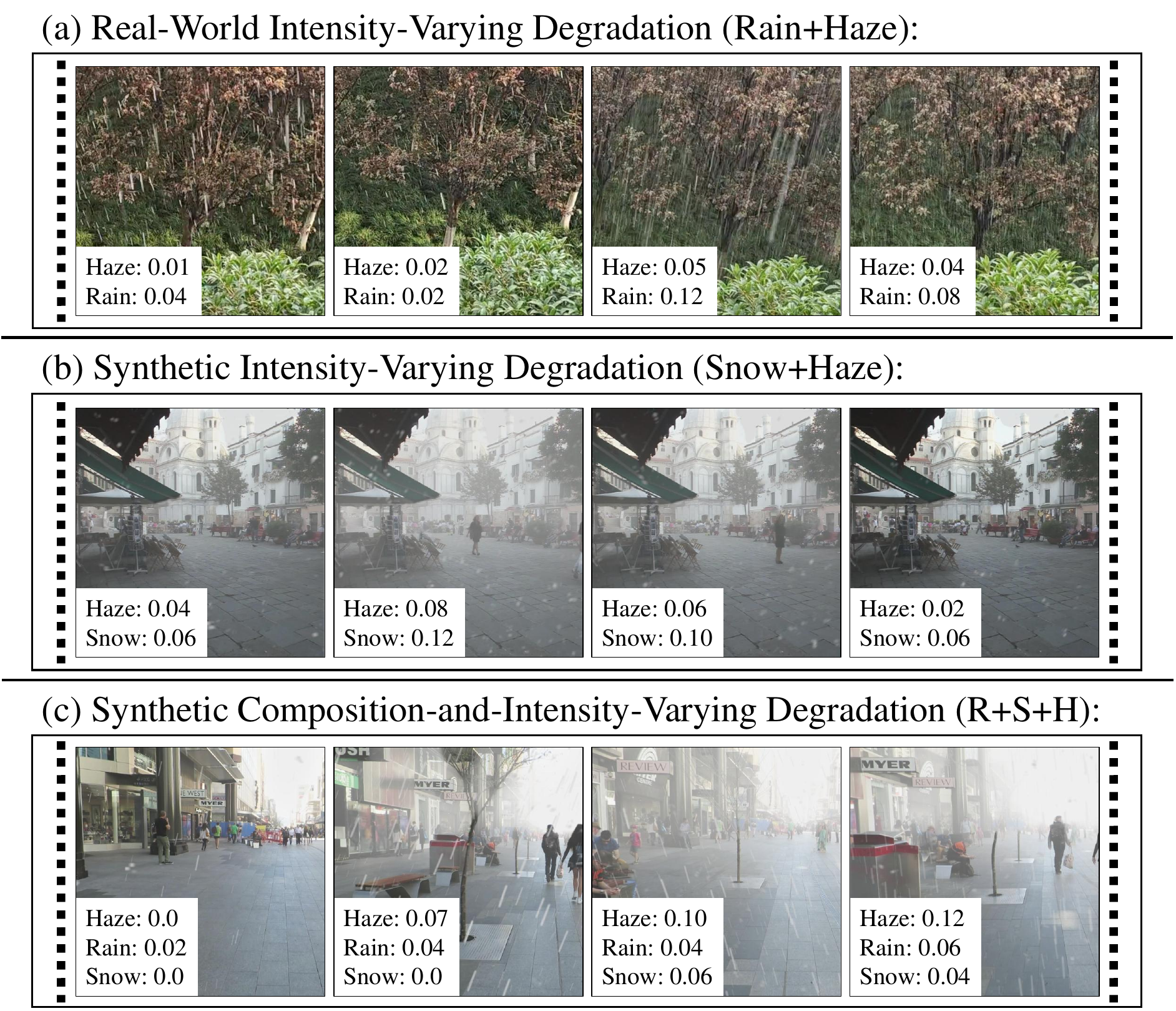}
    \caption{Examples of SEUD scenario. (a) is a real-world SEUD scenario captured by smartphone. (b) and (c) are synthetically generated sequences. In (c), the set of present degradation types changes over time in addition to intensity variation. The numerical values below each frame represent the estimated normalized degradation intensities used in synthesis process.}
    \label{fig:intro}
\end{figure}

With the increasing demand for video processing, AiOIR is gradually extending from images to videos. Video restoration is more challenging than image restoration because it must capture both spatial and temporal dependencies \cite{basicvsrplus2021kelvin, RVRT2022Liang}. Moreover, in video all-in-one restoration, degradations are not only spatially complex but also temporally variant. These temporal variations introduce additional challenges.
Existing studies have made preliminary research in this direction. Zhao et al. introduce the Time-varying Unknown Degradation (TUD) scenario \cite{aver2024zhao}, which extends the all-in-one restoration problem from static images to temporally dynamic videos.
In this setting, both the degradation type and its severity can change across frames, reflecting more realistic and challenging conditions. To construct such a dataset, the TUD framework independently apply degradations to each frame of a clean video. The degradations are randomly sampled from a set including noise, blur, and compression. This design enables comprehensive evaluation of all-in-one models under dynamic and uncertain degradation conditions.

However, there are three limitations in previous work on all-in-one restoration for videos.
(1) \textbf{Limited modeling of temporal correlation of degradation in TUD.} In practice, factors such as rain density, haze thickness, and snow motion change smoothly over time, and different weather types may coexist or transition gradually. In contrast, the TUD setting \cite{aver2024zhao} applies degradations to each frame independently without temporal linkage. This design neglects the physical continuity of real-world degradations. This design breaks the natural continuity of degradation and produces unrealistic temporal changes.
(2) \textbf{Limited modeling of cross-frame dependencies in prompt design.} Prompt-based AiOIR methods mainly focus on single-image restoration. Textual and multimodal prompts require precise external inputs and face cross-modal alignment issues and model complexity \cite{textp2024Marcos, textp2024Hao, textp2025Yan, mulmodel2024Ai, T3diff2024chen}. Visual prompts extract embeddings from the degraded image; but they are usually generated per frame and remain independent across time \cite{PIPli2023, PIRpotlapalli2023, prores2023Ma}. As a result, they fail to capture how degradations evolve. 
(3) \textbf{Limited modeling of compound degradations.} Many all-in-one restoration methods struggle to handle co-occurring degradations. 
For example, PromptIR \cite{PIRpotlapalli2023} and PIP \cite{PIPli2023} do not include mechanisms to prevent similarity or conflict among task prompts. T3DiffWeather \cite{T3diff2024chen} uses a predefined prompt pool, which limits coverage of unseen or mixed degradations. TAP \cite{tap2025Wang} models task relations with low-rank and contrastive constraints but relies on predefined degradation correlations, which limits flexibility in complex or dynamic scenarios.

\begin{figure*}
    \centering
    \includegraphics[width=1\linewidth]{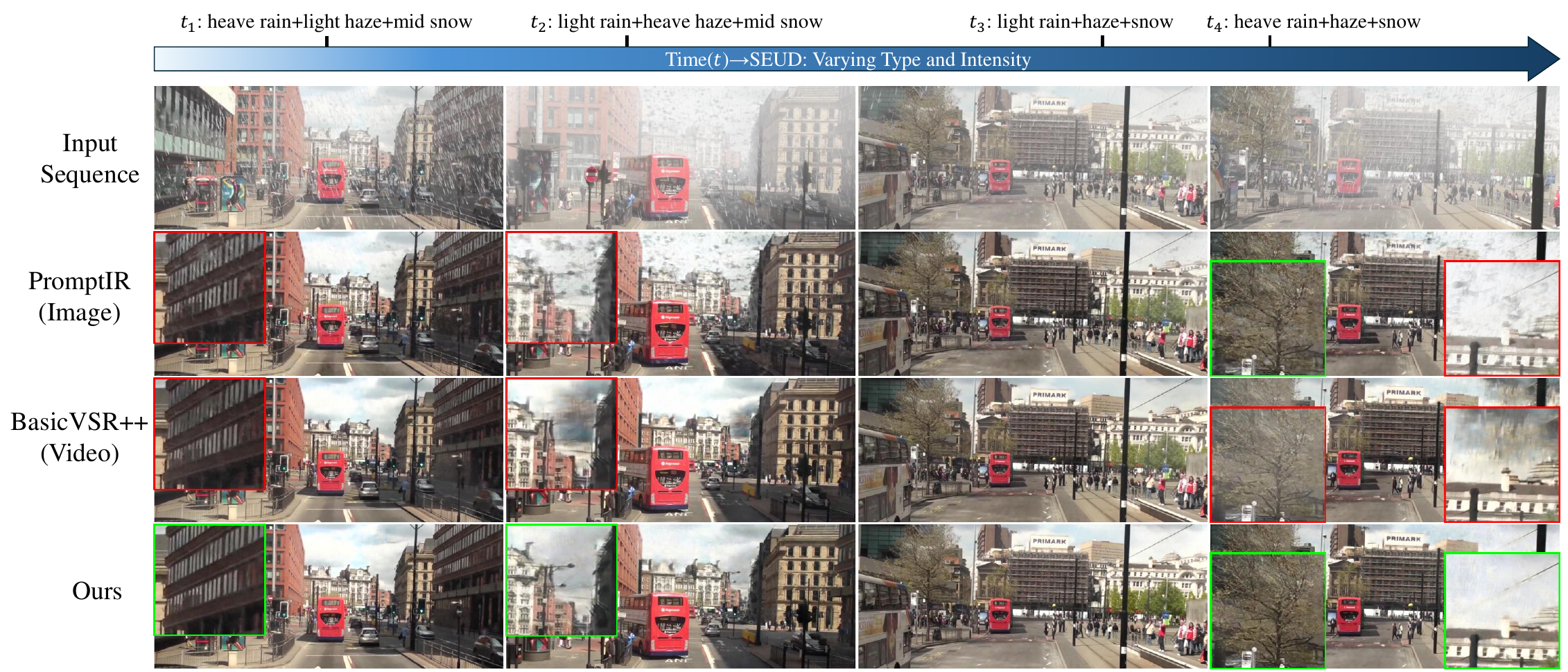}
    \caption{Comparison of recovery results in SEUD scenarios. Existing restoration methods show artifacts and instability under SEUD, while our approach maintains more consistent recovery.}
    \label{fig:introduction3crop}
\end{figure*}

To address the aforementioned limitations, we introduce a new Smoothly Evolving Unknown Degradations (SEUD) scenario. In SEUD, the active compound degradation types and severity both evolve smoothly over time. In this work, we focus on weather degradations within this scenario. Figure~\ref{fig:intro} shows three cases: (a) a real case with a fixed type set and varying intensity (rain+haze), (b) a synthetic case with a fixed type set (snow+haze) and (c) a synthetic case where both the type set and the intensity vary (rain+snow+haze). In every frame, parameters such as rain density, transparency, angle, and streak dimensions change smoothly. This setting better matches real processes and provides a more challenging benchmark for all-in-one video restoration. Figure~\ref{fig:introduction3crop} further shows that representative restoration models struggle under SEUD, exhibiting residual artifacts and inconsistent recovery when degradation type and intensity evolve over time.


Because real SEUD data are difficult to collect, we design a flexible weather synthesis pipeline that takes a clean video and a depth map as inputs. For rain and snow, we generalize static image overlay models to a dynamic particle model. Each particle has physical attributes, including depth position, fall speed, transparency, and other relevant properties. The attribute of depth would control occlusion between foreground and background. Other attributes, such as length, thickness, speed, and transparency, are also correlated with depth. Specifically, distant particles are thinner, shorter, slower, and more transparent, and they are more easily occluded by the foreground. We then define several smooth intensity trajectories over time to modulate each degradation. A video may include segments with no degradation, a single degradation, or compound degradations. The pipeline enforces frame-to-frame coherence while maintaining diversity.

In this paper, we propose a novel ORCANet for the SEUD scenario. 
As shown in Fig.~\ref{fig:introduction3crop}, our method produces more uniform and stable restoration across evolving degradations compared with existing approaches.
Specifically, ORCANet consists of two key components: a Coarse Intensity Estimation Dehazing (CIED) module and a Flow Prompt Generation (FPG) module. CIED first estimates haze intensity coarsely and applies a physics-inspired dehazing before feature propagation. 
This step mitigates the multiplicative attenuation of haze that masks other degradations.
FPG extends static visual prompts from AiOIR to temporal prompts for video. It generates static and dynamic prompts. Specifically, the static prompt is shared across multiple frames. It is computed on key frames every $p$ frames and reused for intermediate frames. The static prompt serves as a segment-level condition that separates degradation types and remains stable within short temporal windows. We supervise static prompt with a label-aware metric loss, which pulls descriptors of similar types together and pushes dissimilar ones apart. This supervision mitigates prompt similarity or conflict and improves perception of compound degradations. The dynamic prompt varies with each frame. It propagates with features and fuses with the newly generated dynamic prompt at the current frame to maintain temporal linkage. 
Finally, the static and dynamic prompts are multiplied, and the fused prompt guides degradation perception and adaptive restoration.
Extensive experiments on synthetic and real data show that ORCANet achieves strong performance on SEUD and outperforms existing methods.
The main contributions of this work are as follows:
\begin{itemize}
    \item We introduce Smoothly Evolving Unknown Degradations (SEUD) scenario, which incorporats temporal continuity in both composition and intensity variation, better matching real-world conditions.

    \item We propose a simple yet flexible pipeline for generating SEUD video sequences, capable of modeling the realistic temporal evolution of multiple degradations. We release the synthetic code and dataset to facilitate benchmarking.
    
    \item We develop ORCANet, an all-in-one video restoration model that extends prompt-based degradation modeling from images to videos, enabling adaptive restoration under smoothly evolving degradations for SEUD scenario.



    \item Extensive experiments demonstrate that our approach achieves superior performance on both synthetic and real-world SEUD videos.
\end{itemize}

\section{Related work}
\subsection{Weather Degradation Models}
Early studies in adverse weather restoration often rely on physical models that describe how degradations form under different weather conditions. These formulations provide interpretable priors for image restoration. 

For hazy scenes, the Koschmieder’s atmospheric scattering model \cite{fogsim1959isra} is widely used. It assumes that the observed image $L(x)$ is a combination of clear image $H(x)$ attenuated by the transmission map $t(x)$ and the global atmospheric light $A_\infty$:
\begin{equation}
L(x) = H(x) \cdot t(x) + A_{\infty} \cdot (1 - t(x)),
\label{eq:haze}
\end{equation}
where the transmission $t(x) = e^{-\beta d(x)}$ depends on the scattering coefficient $\beta$ and the scene depth $d(x)$. Eq.~\eqref{eq:haze} explains two key haze effects: contrast loss through multiplicative attenuation $t(x)$ and color shifting through the additive airlight term. Many classical and modern dehazing methods estimate $t(x)$ and $A_\infty$ from a single image, sometimes with priors, learning-based estimators, or depth surrogates \cite{yoly2021you, Ye2022ECCVDensity, Song2022VITDehaze, D42022yang, BMVC2024Kirillova}.

Rain and snow introduce a semi-transparent precipitation layer that occludes or blends with the scene. A unified and practical model uses per-pixel alpha compositing:
\begin{equation}
L(x) = H(x) \cdot (1 - \alpha(x)) + R(x) \cdot \alpha(x),
\end{equation}
where $R(x)$ denotes the rain or snow layer, and  $\alpha(x)\!\in\![0,1]$ is a transparency map that controls the visibility of degradations. In rain, $R(x)$ often contains thin, oriented, motion-blurred streaks. In snow, $R(x)$ contains larger, blob-like, or out-of-focus particles.
Many restoration methods exploit this layered view and design priors or networks for $R(x)$ and $\alpha$ \cite{Qian2018Raindrop, Quan2021OneGo, Chen2020JSTASR}.

These degradation models offer simplified yet effective representations of adverse weather phenomena. They provide strong physical priors that guide the design of weather-specific restoration methods, such as dehazing \cite{Ye2022ECCVDensity, Song2022VITDehaze, D42022yang, BMVC2024Kirillova, UCL24wang}, deraining \cite{Qian2018Raindrop, sid2021yang, ad2016you, hearain2019li}, and desnowing \cite{desnow2018liu, Quan2021OneGo, Chen2020JSTASR, deepse2021zhang, dis2023quan}. While these models enable effective restoration under static and single-frame assumptions, they do not explicitly model the temporal continuity of weather degradations. In practice, both degradation composition and intensity evolve smoothly over time, which limits the applicability of these formulations in dynamic video scenarios.

\begin{figure*}
    \centering
    \includegraphics[width=1\linewidth]{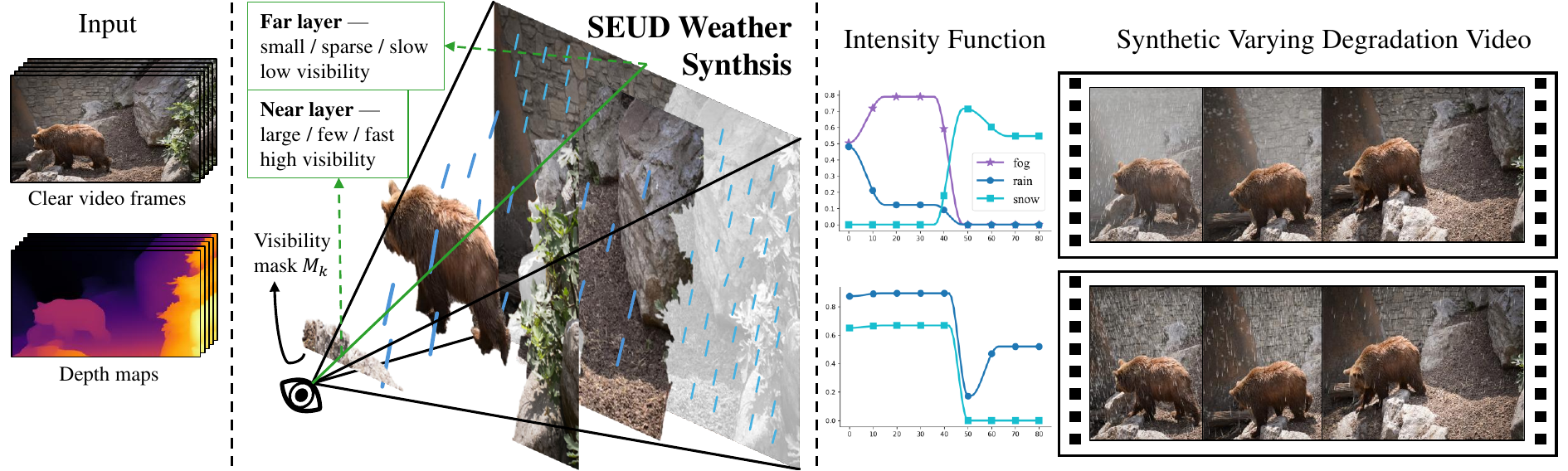}
    \caption{SEUD Weather Synthesis Pipeline. 
    \textbf{Left:} The input consists of clear video frames and their corresponding depth maps. 
    \textbf{Middle:} 
    Each particle is assigned a random depth and a set of physical attributes which correlate with depth. 
    Distant particles are thinner, shorter, slower, and more transparent, and are more likely to be occluded by the foreground visibility mask. Their motion follows gravity $\mathbf{g}$ and the time-varying wind field $\mathbf{w}(t)$. 
    \textbf{Right:} Example intensity functions $\beta(t)$ and $f(t)$ control the temporal evolution of haze and precipitation, producing synthetic videos with continuous, mixed, and time-varying degradations.
    }
    \label{fig:synthsispip}
\end{figure*}

\subsection{All-in-one Restoration}
The goal of all-in-one image restoration is to recover high-quality clean images from low-quality degraded inputs using a single unified network. In the field of weather degradation removal, the pioneering work All-in-One \cite{aiowea2020Li, transweather2022Jose, air2022Li, PIPli2023, wdiffusion2023ozan, PerIR25zhang, UniUIR25zhang, relation25Li} inspires a series of unified models. TransWeather \cite{transweather2022Jose} and TKL \cite{TKL2022chen} design specific encoder-decoder structures to handle different weather degradations. AWRCP \cite{AWRCP2023Ye} introduces a high-quality codebook prior and employs a pretrained VQGAN to recover fine texture details under adverse weather. Zhu et al. \cite{twostage2023zhu} present a two-stage training strategy to improve adaptability across various weather conditions. More recently, T3DiffWeather \cite{T3diff2024chen} proposes a prompt pool mechanism, where the network selects the most relevant prompts based on image features for targeted restoration.
Research further extends from image-level restoration to video-level scenarios. ViWS-Net \cite{ViWS2023Yang} introduces temporally-active messenger tokens to collect weather-related information during video encoding. Diff-TTA \cite{difftsc2024yang} combines diffusion models with test-time adaptation to remove adverse weather degradations under both known and unknown conditions. Xu et al. \cite{visl2025xu} provide pseudo-labels for real photos using various large visual language models and restore images with a semi-supervised framework. CUDN \cite{cudn2023Cheng} proposes a sequence-adaptive degradation estimator for predicting the degradation matrix of the entire video sequence. 

However, video all-in-one restoration is still in an early stage. Compared with image tasks, video restoration faces the key challenge that degradations often change dynamically over time. Zhao et al. \cite{aver2024zhao} discuss time-varying unknown degradations (TUD) in earlier all-in-one work. Yet, their method assumes that degradations are independent across frames. In real-world scenarios, degradations are usually continuous and correlated between adjacent frames. This work therefore focuses on the problem of smoothly evolving weather degradations in videos and explores effective modeling.

\subsection{Prompt Learning for Restoration}
Prompt learning has recently gained popularity in all-in-one image restoration, as it provides additional guidance for handling diverse degradations. Prompts can be categorized into visual prompts \cite{PIPli2023, PIRpotlapalli2023, prores2023Ma, MaIR2025Li, T3diff2024chen, tap2025Wang}, textual prompts \cite{textp2024Marcos, textp2024Hao, textp2025Yan}, and multimodal prompts \cite{mulmodel2024Ai, T3diff2024chen}. Textual and multimodal prompts require precise text inputs as external guidance, but face challenges of cross-modal alignment and large-scale language model complexity. Visual prompts instead extract embeddings from degraded images to guide the network. However, they may suffer from parameter conflicts across tasks. Among visual prompt methods, PromptIR \cite{PIRpotlapalli2023} introduces a lightweight prompt block to enhance all-in-one image restoration. ProRes \cite{prores2023Ma} integrates target visual prompts into input images and employs weighted combinations for customized restoration. PIP \cite{PIPli2023} proposes a Prompt-In-Prompt framework that restores clean images relying only on the input without prior knowledge of degradation types. More recently, T3DiffWeather \cite{T3diff2024chen} develops a prompt pool mechanism that adaptively selects the most relevant prompts according to image features at inference time. TAP \cite{tap2025Wang} proposes a parameter-efficient task-aware prompting framework that models inter-task relationships through low-rank decomposition and contrastive constraints to mitigate task conflicts in all-in-one adverse weather image restoration.

Most existing methods design prompts for image restoration. Although these approaches can be applied to frame-wise video restoration, the generated prompts are usually independent across frames. They fail to capture temporal dependencies in video degradations. In the task of SEUD, degradations across frames exhibit both shared patterns (e.g., consistent weather type within a short period) and unique variations (e.g., intensity or wind direction). To address this problem, we propose to divide prompts into two components: The first component captures temporally invariant degradation cues that are shared among nearby frames, while the second models frame-specific variations that evolve smoothly over time. This design allows the model to exploit both temporal consistency and local dynamic changes. By combining the two, the network learns temporally coherent yet adaptive representations for video restoration under SEUD conditions.

\section{The proposed method}
In this section, we first present the physical models of time-varying weather degradation for video in Section~\ref{sec:video_weather}.
Sections~\ref{sec:overview_net}--~\ref{sec:FPG} then describe the overall framework of our method, which integrates an intensity-aware mechanism, an all-in-one visual prompting design for video, and a unified restoration network.
Finally, Section~\ref{sec:trainstra} details the design of loss functions and training strategies.

\subsection{Time-Varying Weather Degradation for Video}\label{sec:video_weather}

The adverse weather in the video exhibits continuous and time-varying behavior. The intensity evolves smoothly or in segments over time. Different weather types may overlap on the temporal axis and transition into each other. To address the difficulty of obtaining real-world data, we propose a practical data synthesis method. We adopt a unified view that combines volumetric scattering and foreground particle occlusion. We model haze, rain, and snow in space and time, and provide a matching synthesis pipeline. 
It explicitly encodes the temporal evolution of intensity, which supports robust generalization under continuous, blind, and mixed degradations. The simple synthesis process is illustrated in Fig.~\ref{fig:synthsispip}.

\paragraph{Video Haze Model}
Video haze follows a volumetric scattering formulation. Let $H(x,t)\in \mathbb{R}^{H\times W \times 3}$ denote the clean video, where $x$ is the pixel location and $t$ is time. We extend Eq.~\eqref{eq:haze} to the video domain and adopt a time-varying transmission $T(x,t)$ in the \cite{fogsim1959isra}:
\begin{equation}
L(x,t) = H(x,t)\,T(x,t) + A_\infty\!\left(1 - T(x,t)\right),
\label{eq:video_haze}
\end{equation}
where $A_\infty$ is the atmospheric light, which is approximately constant within a short time window. 
The transmission term $T(x,t)$ represents the proportion of scene radiance that reaches the camera after scattering attenuation, and it depends jointly on the scene depth $D(x)$ and a scattering coefficient $\beta(t)$:
\begin{equation}
T(x,t) = \exp\!\big(-\beta(t)\,D(x)\big),
\end{equation}
where the coefficient $\beta(t)$ describes the temporal variation of haze density, which can evolve smoothly or in segments over time.
Eq.~\eqref{eq:video_haze} captures both the depth-dependent contrast attenuation and the color shift due to airlight. Camera motion and slight illumination changes do not affect the physical relation but only affect $H(x,t)$.

\paragraph{Video Rain and Snow Model}
Precipitation introduces a semi-transparent foreground layer between the camera and the scene. We model it at the particle level and use alpha compositing for occlusion and blending. Let $Z_{\rm scene}(x)$ be the scene depth at pixel $x$. For the $k$-th particle at time $t$, let $Z_k(x,t)$ be its depth along the viewing ray and $P_k(x,t)$ its radiance in the image. The visibility mask is formulated as:
\begin{equation}
M_k(x,t) = \mathbf{1}\!\left\{\, Z_k(x,t) < Z_{\rm scene}(x) \,\right\}.
\end{equation}
We composite particles in a forward, per-particle manner:
\begin{equation}
\begin{aligned}
L_{k+1}(x,t)
&= \big(1-\alpha_k(x,t)\,M_k(x,t)\big)\,L_k(x,t)\\
&+ \, \alpha_k(x,t)\,M_k(x,t)\,P_k(x,t),
\end{aligned}
\label{eq:alpha_blend}
\end{equation}
where $L_0(x,t)$ is the hazy frame given by Eq.~\eqref{eq:video_haze}. The term $\alpha_k(x,t)\!\in[0,1]$ is the effective opacity that aggregates particle optical thickness, size, and local imaging. In practice, $P_k(x,t)$ is implemented by a texture sprite or a small convolutional kernel. Swapping the texture or kernel produces different particle appearances. Rain and snow differ only in their dynamics and optical characteristics.
Rain produces thin, oriented, and motion-blurred streaks with higher velocity,
whereas snow consists of larger, slower, and often out-of-focus blobs.

\begin{figure*}
    \centering
    \includegraphics[width=1\linewidth]{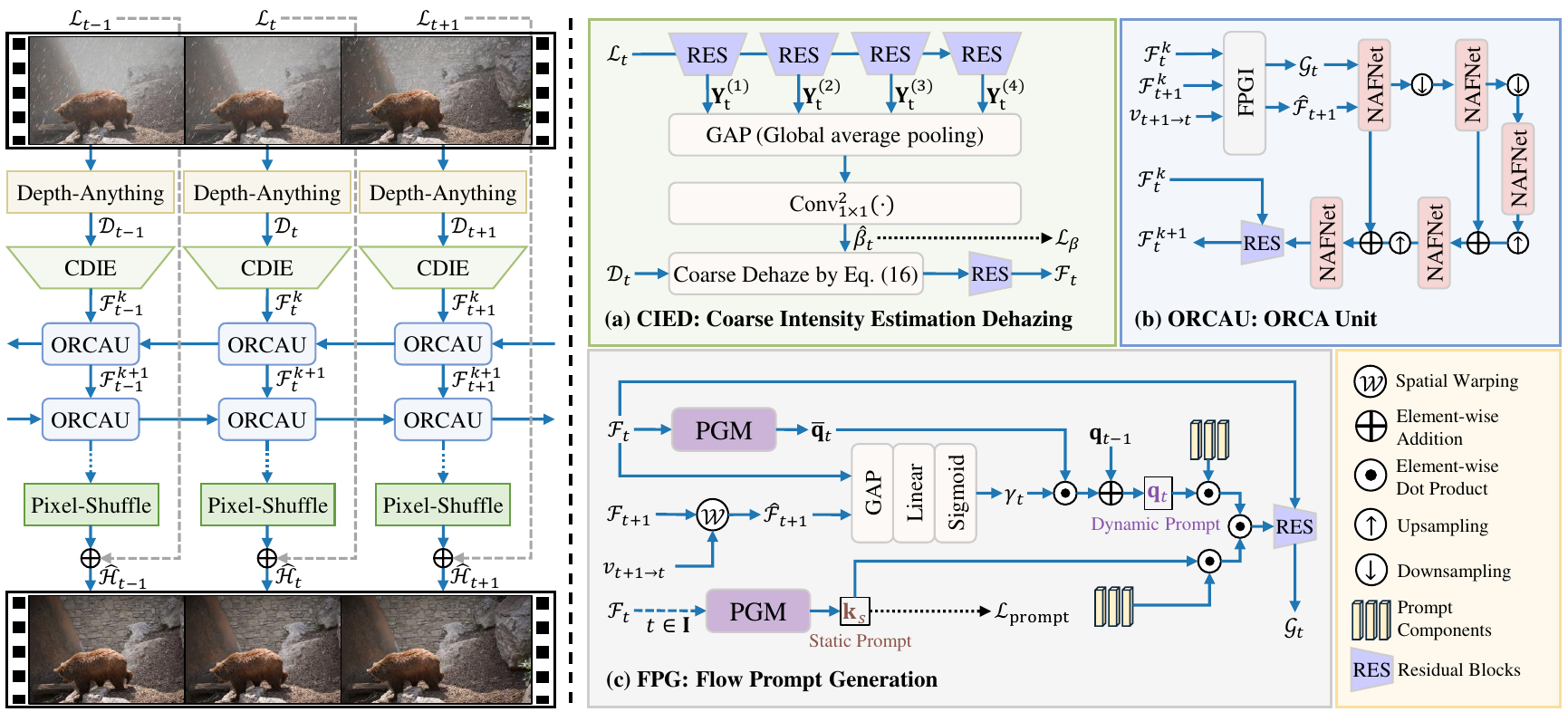}
    \caption{\textbf{OCRANet Framework Overview.} 
    The network adopts a bidirectional recurrent propagation design for video restoration. 
    Each frame is first processed by the CIED module for depth-guided coarse dehazing and intensity estimation. 
    Temporal features are refined through multiple ORCA Units (ORCAU) with flow-aligned prompts generated by the FPG module. 
    A pixel-shuffle decoder reconstructs clean frames. 
    Detailed module structures of CIED, ORCAU, and FPG are shown on the right (a)-(c).}
    \label{fig:mainflow}
\end{figure*}

We update the image-plane motion of each particle by
\begin{equation}
\mathbf{u}_k(t{+}1) = \mathbf{u}_k(t) + \big(\,\mathbf{g}(Z_k)+\mathbf{w}(t)\,\big),
\end{equation}
where $\mathbf{u}_k(t)$ denotes the image-plane velocity of the $k$-th particle at time $t$. The term $\mathbf{g}(Z_k)$ models gravity and perspective effects that depend on particle depth $Z_k$, which cause particles closer to the camera to move faster in the image. The term $\mathbf{w}(t)$ represents a time-varying wind field shared by all particles.
Particles are removed once they leave the image or reach the end of their lifetime. At each time step, new particles are generated according to the current weather intensity. We describe the temporal evolution of rain or snow intensity using a density function $f(t)$, which specifies the expected number of particles per unit image area at time $t$. The function $f(t)$ follows several analytic forms that capture common weather evolution patterns, including step, trapezoid, Gaussian, and cosine profiles.
These temporal functions jointly control both the haze scattering coefficient $\beta(t)$ and the particle density $f(t)$, ensuring consistent evolution across different degradation types. The particle-based synthesis process and the corresponding temporal intensity functions are illustrated in the middle and right parts of Fig.~\ref{fig:synthsispip}.

\

The above model synthesizes continuous, blind, and mixed degradations with a set of interpretable temporal functions and particle parameters. The pipeline is simple to implement and easy to extend, and we will release the corresponding code.

\subsection{Network Overview}\label{sec:overview_net}
We propose an all-in-One Recurrent Conditional and Adaptive prompting Network OCRANet, a unified video restoration network for SEUD task. The overall framework is illustrated in Fig.~\ref{fig:mainflow}. The model follows a recurrent temporal propagation paradigm and operates end-to-end on blind weather-degraded videos of arbitrary length. Given a degraded sequence, the network alternates bidirectional propagation in the time domain. At the $k$-th iteration, if $k$ is odd, features propagate backward ($T\!\rightarrow\!1$); if $k$ is even, features propagate forward ($1\!\rightarrow\!T$). This schedule leverages complementary context without explicit extra cost. It improves long-range dependency aggregation and stabilizes estimates near occlusion boundaries. To strengthen geometric and motion constraints, the framework includes lightweight depth and optical flow branches (initialized from Depth-Anything \cite{depany2024yang} and SPyNet \cite{Spy2017Anurag}). These branches are trained jointly with the main restoration network, which lets the model balance external priors and data fidelity during learning. After the two-way propagation, the decoder fuses the terminal features and reconstructs a clean video via residual blocks and a pixel-shuffle \cite{pixshuf2016shi} head. 

Formally, the input is a $T$-frame degraded sequence $\mathcal{L}=\{L(x,t)\in\mathbb{R}^{3\times H\times W}\}_{t=1}^{T}$. We first extract per-frame depth features using Depth-Anything, yielding $\mathcal{D}=\{D(x,t)\in\mathbb{R}^{C\times H\times W}\}_{t=1}^{T}$. For brevity, we use $\mathcal{L}_t$ and $\mathcal{D}_t$ to denote the $t$-th frame in $\mathcal{L}$ and $\mathcal{D}$, respectively.  We adopt the same shorthand for other frame-indexed sequences in the rest of the paper. We then feed $(\mathcal{L}_t,\mathcal{D}_t)$ into a Coarse Intensity Estimation Dehazing (CIED) module to perform physics-guided coarse dehazing and to produce shallow features:
\begin{equation}
(\mathcal{F}_t,\hat{\beta}_t)=\operatorname{CIED}(\mathcal{L}_t,\mathcal{D}_t),\qquad \mathcal{F}_t\in\mathbb{R}^{C\times H\times W},
\end{equation}
where $\hat{\beta}_t$ is the estimated haze strength at time $t$, and $\mathcal{F}_t$ serves as the shallow feature for subsequent propagation.

Next, we estimate optical flow $\mathcal{V}=\{v_{t+1\rightarrow t},\,v_{t\rightarrow t+1}\}_{t=2}^{T}$ between frames using SPyNet \cite{Spy2017Anurag}. For $k$-th backward propagation, let $\mathcal{F}^{k+1}_{t+1}$ be the hidden propagation feature at time $t+1$. We align it to time $t$ via a warping operator $\mathcal{W}(\cdot)$:
\begin{equation}
\hat{\mathcal{F}}^k_{t+1}=\mathcal{W}\!\big(\mathcal{F}^{k+1}_{t+1},\,v_{t+1\rightarrow t}\big).
\end{equation}

To adapt the representation to the current degradation pattern under the alignment prior, we introduce a Flow Prompt Generation (FPG) module. Given $(\mathcal{F}^k_t,\hat{\mathcal{F}}^k_{t+1})$, FPG produces and injects a visual prompt that modulates the feature according to the degradation state:
\begin{equation}
\mathcal{G}^k_t=\operatorname{FPG}(\mathcal{F}^k_t,\,\hat{\mathcal{F}}^k_{t+1}).
\end{equation}

We then feed $\mathcal{G}^k_t$ and $\hat{\mathcal{F}}^k_{t-1}$ to the restoration backbone to compute the current propagation feature:
\begin{equation}
\mathcal{F}^{k+1}_{t}=\operatorname{Backbone}(\mathcal{G}^k_t,\,\hat{\mathcal{F}}^k_{t+1}).
\end{equation}

We adopt NAFNet \cite{simplebaselinesimagerestoration2022chen} as the backbone in this work due to its efficiency and strong representation with channel attention and point-wise gating. It fits the recurrent propagation well by reducing parameters and FLOPs while preserving fidelity. The $k$-th forward pass mirrors the backward pass by aligning $\mathcal{F}^k_{t-1}$ with $v_{t-1\rightarrow t}$ and updating $\mathcal{F}^k_{t}$.

After a preset number of alternating forward and backward iterations, the decoder fuses the terminal features from both directions. Then it applies several residual blocks and a pixel-shuffle head to produce the final reconstructions $\hat{\mathcal{H}}=\{\hat{H}(x,t)\in\mathbb{R}^{3\times H\times W}\}_{t=1}^{T}$. The parameters of the depth and flow branches are counted in the total model size and are optimized jointly with the backbone and FPG under a unified objective. This joint training allows geometric alignment, physical intensity estimation, and appearance restoration to converge coherently within one framework.

In summary, OCRANet integrates depth estimation, flow-guided alignment, and prompt-driven propagation in a bidirectional recurrent design to address time-varying degradations in a unified way. The pipeline follows a coarse-to-fine strategy: physics-guided initialization via CIED, followed by feature-level refinement through recurrent aggregation. The bidirectional schedule enforces temporal consistency and enhances detail compensation, while FPG improves sensitivity to multiple degradation types and their continuous evolution. The next sections detail each module and training losses. The detailed module designs, including CIED and FPG, are shown in the right part of Fig.~\ref{fig:mainflow} and described in the following subsections.

\subsection{Coarse Intensity Estimation Dehazing (CIED)}\label{sec:CIED}

CIED provides a physics-guided initialization that estimates haze intensity and performs a coarse inverse dehazing before recurrent propagation. Given a degraded frame $L(x,t)$ and its depth map $D(x,t)$, the module predicts a scalar haze strength $\hat{\beta}_t$ per frame, computes the transmission $T(x,t)=\exp\!\big(-\hat{\beta}_t\,D(x,t)\big)$, estimates the atmospheric light $A_\infty$, and applies the inverse of Eq.~\eqref{eq:video_haze} to obtain a coarse clean image. We extract multi-scale features from serious residual blocks. Let $\{\mathbf{Y}^{(s)}_t\}_{s=1}^4$ denote the features at four stages and $\operatorname{GAP}(\cdot)$ be global average pooling. CIED forms a compact descriptor by channel-wise pooling and concatenation, passes it through two $1{\times}1$ convolutions, and regresses an unnormalized intensity $\tilde{\beta}_t$:
\begin{align}
&[\mathbf{z}^{(1)}_t,\mathbf{z}^{(2)}_t,\mathbf{z}^{(3)}_t,\mathbf{z}^{(4)}_t]
 = [\operatorname{GAP}(\mathbf{Y}^{(1)}_t),\dots,\operatorname{GAP}(\mathbf{Y}^{(4)}_t)],\\
&\tilde{\beta}_t
 = \operatorname{Conv}_{1\times 1}^2\!\big(\operatorname{Concat}(\mathbf{z}^{(1)}_t,\dots,\mathbf{z}^{(4)}_t)\big),
\end{align}
where $\operatorname{Concat}(\cdot, \cdot)$ denotes channel-wise concatenation. We squash $\tilde{\beta}_t$ with a $\tanh$ and map it to a physically plausible interval $[\beta_{\min},\beta_{\max}]$:
\begin{equation}
\hat{\beta}_t = \beta_{\min} + \frac{\beta_{\max}-\beta_{\min}}{2}\,\big(1+\tanh(\tilde{\beta}_t)\big).
\end{equation}

CIED then computes the transmission field:
\begin{equation}
\hat{T}(x,t) = \exp\!\big(-\hat{\beta}_t\,D(x,t)\big).
\end{equation}

The atmospheric light $A_\infty$ is estimated from the input by channel-wise spatial maximum. With $\hat{T}$ and $A_\infty$, CIED applies the inverse of the forward haze model in Eq.~\eqref{eq:video_haze} to produce a coarse dehazed image $\tilde{H}(x,t)$:
\begin{equation}
\tilde{H}(x,t) = \big(L(x,t)-A_\infty\big)\,/\!\max\!\big(\hat{T}(x,t),\epsilon\big) + A_\infty,
\end{equation}
where $\epsilon$ is a small constant for numerical stability. During training on synthetic videos, CIED uses ground-truth intensity labels $\beta_t$ and minimizes mean squared error on scalar estimate:
\begin{equation}
\mathcal{L}_{\beta}=\frac{1}{T}\sum_{t=1}^{T}\big\|\hat{\beta}_t-\beta_t\big\|_2^2.
\label{eq:l_beta}
\end{equation}
This loss anchors the global attenuation level and reduces ambiguity between depth and scattering in $\hat{T}$. Subsequently, a set of residual blocks extract features from coarse dehazed image $\tilde{H}(x,t)$ to obtain $\mathcal{F}_t\in\mathbb{R}^{C\times H\times W}$. The coarse inverse dehazing $\tilde{H}(x,t)$ serves two roles. It provides a perceptual measure of frame-wise haze intensity variation across the video and it supplies shallow features with improved contrast for motion estimation and prompt generation. 

\subsection{Flow Prompt Generation (FPG)}\label{sec:FPG}

FPG generates a degradation-aware visual prompt that conditions the backbone with a slowly varying, type-specific component and a frame-adaptive component, followed by implicit enhancement interaction. Two Prompt Generation Modules (PGM) from PromptIR \cite{PIPli2023} share the same design but serve different roles. The first PGM generates a static prompt descriptor when $t$ is a multiple of $p$. Let $s\in \mathbf{I}=\{k*p|k\in \mathbb{Z} \}$ denote the current segment index. The static prompt descriptor $\mathbf{k}_s\in\mathbb{R}^{d}$ characterizes the degradation type for the segment and remains shared within the next $p$ frames. It separates weather-like degradations and persists as the segment-level condition. When synthetic degradation-type labels are available, we supervise the static prompt descriptors to encode label proximity. Let the multi-hot label of segment $s$ be $\mathbf{y}_s\in\{0,1\}^{K}$. We define a cosine label affinity:
\begin{equation}
a_{ij}=\frac{\mathbf{y}_i^\top\mathbf{y}_j}{\|\mathbf{y}_i\|_2\|\mathbf{y}_j\|_2}\in[0,1].
\end{equation}
We optimize a label-aware metric loss that pulls descriptors from similar types together and pushes dissimilar ones apart:
\begin{equation}
\begin{aligned}
\mathcal{L}_{\rm prompt}
=\sum_{i}&\Bigg[
\frac{1}{|\mathcal{P}(i)|}\sum_{j\in\mathcal{P}(i)}\big\|\bar{\mathbf{k}}_i-\bar{\mathbf{k}}_j\big\|_2^2 \\
&+\frac{1}{|\mathcal{N}(i)|}\sum_{j\in\mathcal{N}(i)}\big[m-\big\|\bar{\mathbf{k}}_i-\bar{\mathbf{k}}_j\big\|_2\big]_+^2
\Bigg],
\end{aligned}
\label{eq:l_basis}
\end{equation}
where $\bar{\mathbf{k}}_s = \mathbf{k}_s / \|\mathbf{k}_s\|_2$, $\mathcal{P}(i)=\{j\,|\,a_{ij}\ge\tau_{+}\}$, $\mathcal{N}(i)=\{j\,|\,a_{ij}\le\tau_{-}\}$ with $\tau_{+}>\tau_{-}$, $m$ is a margin, and $[\cdot]_+$ denotes the hinge operator.

The second PGM generates an intermediate feature $\bar{\mathbf{q}}_t$ for each frame based on the shallow feature $\mathcal{F}_t$ and the flow-aligned hidden state $\hat{\mathcal{F}}_{t+1}$. This feature is further refined through a gated update with the previous prompt to produce the final dynamic prompt descriptor $\mathbf{q}_t$:
\begin{align}
&{\gamma}_t=\sigma\big(\Gamma(\operatorname{GAP}(\operatorname{Concat}(\mathcal{F}_t,\hat{\mathcal{F}}_{t+1})))\big)\in(0,1)^d,\\
&\mathbf{q}_t={\gamma}_t\odot \bar{\mathbf{q}}_t+ \mathbf{q}_{t-1},
\label{eq:dpd_gate}
\end{align}
where $\Gamma(\cdot)$ is linear projection, $\sigma$ is the sigmoid, and $\odot$ is the Hadamard product.

Given the static prompt $\mathbf{k}_s$ and the dynamic prompt $\mathbf{q}_t$, FPG assigns independent learnable spatial kernel banks to each descriptor: $\mathbf{P}^{(k)}\in\mathbb{R}^{d\times C\times H_0\times W_0}$ for the static prompt path and $\mathbf{P}^{(q)}\in\mathbb{R}^{d\times C\times H_0\times W_0}$ for the dynamic prompt path. Each descriptor forms its spatial prompt via a coefficient-weighted combination over channels:
\begin{equation}
\mathbf{U}^{(k)}_t=\sum_{c=1}^{d}k_{s,c}\mathbf{P}^{(k)}_{c},\qquad
\mathbf{U}^{(q)}_t=\sum_{c=1}^{d}q_{t,c}\mathbf{P}^{(q)}_{c}.
\label{eq:spatial_prompts}
\end{equation}
We then fuse the two spatial prompts by element-wise product:
\begin{equation}
\mathbf{U}_t=\mathbf{U}^{(k)}_t\odot \mathbf{U}^{(q)}_t.
\label{eq:prompt_fusion}
\end{equation}

Since the spatial resolution of $\mathbf{U}_t$ may differ from that of the current feature, we apply bilinear interpolation to match the target size $(H, W)$ before injecting it into the backbone.
Finally, we inject the fused prompt by concatenating it with the current backbone input and applying a residual block:
\begin{equation}
\mathcal{G}_t=\operatorname{Res}\big(\operatorname{Concate}(\mathcal{F}_t,\mathbf{U}_t)\big).
\label{eq:res_inject}
\end{equation}

This design separates the spatial modulation of long-term degradation identity and short-term frame attributes. The Static Prompt is trained with a label-aware loss to distinguish different restoration tasks across segments, while the Dynamic Prompt refines the guidance between adjacent frames. In this way, the model applies prompt-based conditioning to restore continuous, time-varying, and mixed degradations in videos.

\subsection{Training Strategy}\label{sec:trainstra}

We train ORCANet with three complementary objectives: a haze-intensity regression loss $\mathcal{L}_{\beta}$ (Eq.~\eqref{eq:l_beta}), a label-aware basis metric loss $\mathcal{L}_{\rm prompt}$ (Eq.~\eqref{eq:l_basis}), and a video Charbonnier loss \cite{loss1994Char}. 
Let $\rho(x)=\sqrt{x^{2}+\varepsilon^{2}}$ be the Charbonnier penalty with a small constant $\varepsilon>0$. Denote the restored frame by $\hat{\mathcal{H}}_{t}$ and the ground-truth clean frame by $\mathcal{H}_t$. The spatial Charbonnier term promotes per-frame reconstruction fidelity:
\begin{equation}
\mathcal{L}_{\text{char}}
=
\frac{1}{T}\sum_{t=1}^{T}\;\frac{1}{|\Omega|}\sum_{x\in\Omega}\rho\!\big(\mathcal{H}_t(x)-\hat{\mathcal{H}}_{t}(x)\big),
\label{eq:l_char_spatial}
\end{equation}
where $\Omega$ indexes all pixel locations and channels. The final objective is a weighted sum:
\begin{equation}
\mathcal{L}_{\text{total}}
=
\mathcal{L}_{\text{char}}
+
\lambda_{\beta}\;\mathcal{L}_{\beta}
+
\lambda_{\text{prompt}}\;\mathcal{L}_{\rm prompt},
\label{eq:l_total}
\end{equation}
where $\lambda_{\beta}, \lambda_{\text{prompt}}$ balance the terms. We set $\lambda_{\beta}=0.5$ and $\lambda_{\text{prompt}}=0.2$ in all experiments. For the prompt supervision in Eq.~\eqref{eq:l_basis}, we assign $\tau_+ = 0.7$ and $\tau_- = 0.3$, and we set the margin to $m=0.5$. 

For implementation, we use the same configuration across all experiments. The channel width of the network is set to $64$, and the prompt embedding length and spatial size are set to $5$ and $96\times96$, respectively. The pretrained modules, Depth-Anything~\cite{depany2024yang} and SPyNet~\cite{Spy2017Anurag}, are integrated into our framework, and their parameters and runtime are included. During inference, long videos are divided into non-overlapping clips of $80$ frames, which are restored sequentially to ensure stable memory usage and temporal consistency.

All experiments run on a server with an Intel(R) Xeon(R) Platinum 8368 CPU and four NVIDIA A100-SXM4 GPUs with 40\,GB memory each. We train the network with the Adam optimizer~\cite{adam2017kingma} using $\beta_1=0.9$ and $\beta_2=0.999$. The input clip length is $T=12$ frames and the training resolution is $256\times256$. We use a batch size of $1$ and optimize for $600{,}000$ iterations. We jointly update all parameters, including those from pretrained components, under the objective in Eq.~\eqref{eq:l_total}.

\section{Experimental Results}
\subsection{Dataset and Evaluation Metrics}

\begin{table*}[htbp]
\centering
\fontsize{9}{12}\selectfont
\setlength{\tabcolsep}{8pt}
\caption{Comparison on Mot17 and DAVIS-test under degradation setting 1 and setting 2.}
\begin{tabular}{c|cc|cc|cc|cc|cc}
\toprule
\multirow{3}{*}{Method} & 
\multicolumn{4}{c|}{Mot17} & 
\multicolumn{4}{c|}{DAVIS-test} &
\multirow{3}{*}{Time (s)} &
\multirow{3}{*}{Params} \\
\cline{2-9}
 & \multicolumn{2}{c|}{Setting 1} & \multicolumn{2}{c|}{Setting 2} & \multicolumn{2}{c|}{Setting 1} & \multicolumn{2}{c|}{Setting 2} & & \\
\cline{2-9}
 & PSNR & SSIM & PSNR & SSIM & PSNR & SSIM & PSNR & SSIM & & \\
\midrule
Transweather \cite{transweather2022Jose} & 18.73 & 0.5506 & 19.94 & 0.6145 & 20.44 & 0.5672 & 20.64  & 0.5839  & 3.06 & 38.1 M \\
AirNet \cite{air2022Li}   & 23.62 & 0.8900 & 26.56 & 0.9412& 24.65  & 0.9034  & 19.93 & 0.8774 & 12.14 & 8.9 M \\
PromptIR \cite{PIPli2023} & 32.98 & 0.9453 & 29.95 & 0.9483 & 30.46 & 0.9361  & 24.91  & 0.8956  & 9.87 & 35.6 M \\
BasicVSR++ \cite{basicvsrplus2021kelvin} & 36.26 & 0.9793 & 32.43 & 0.9787 &  32.33 &  0.9739 & 25.51 & 0.9527  & 4.73 & 7.4 M \\
RVRT \cite{RVRT2022Liang} & 25.63 & 0.8851 & 23.36 & 0.9040 & 27.71 & 0.9225 & 19.39 & 0.8313  & 14.86 & 13.6 M \\
AverNet \cite{aver2024zhao} & 28.98 & 0.8894 & 27.42 & 0.8582 & 25.46 & 0.8023 & 25.01 & 0.7952 & 6.20 & 41.3 M \\
T3DiffWeather \cite{T3diff2024chen} & 34.84 & 0.9409 & 32.47 & 0.9730 & 27.17 & 0.8814 & 26.65 & 0.8842 & 16.46 & 69.4 M \\
OCRANet (Ours) & \textbf{37.88} & \textbf{0.9807}  & \textbf{35.22} & \textbf{0.9795} & \textbf{32.86} & \textbf{0.9852} & \textbf{28.43} & \textbf{0.9687} & 13.54 & 70.0 M \\
\bottomrule
\end{tabular}
\label{tab:singledynamic}
\end{table*}

\begin{figure*}
    \centering
    \includegraphics[width=1\linewidth]{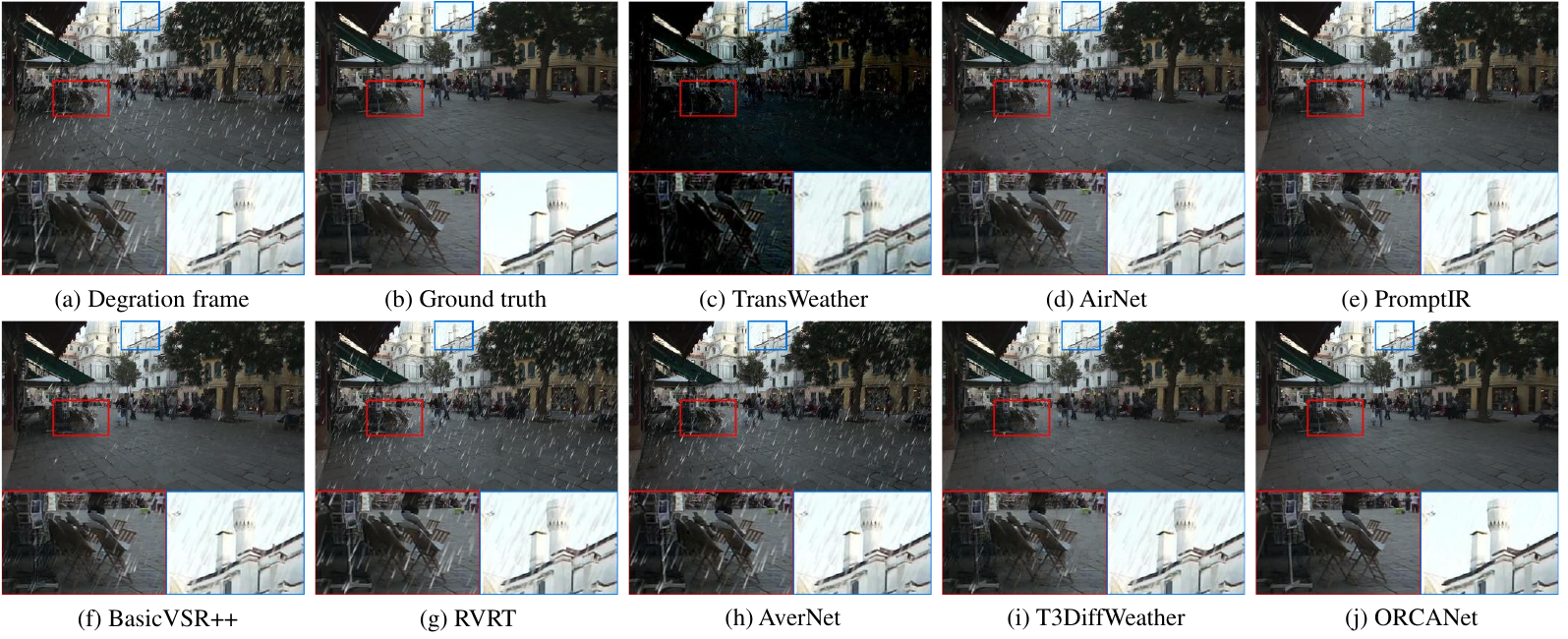}
    \caption{Comparison of visual restoration results of rainy degraded video frame.}
    \label{fig:setting1}
\end{figure*}


We conduct experiments on two widely used video datasets, MOT17 \cite{mot2016milan} and DAVIS \cite{Davis2016F}, to evaluate performance under the SEUD setting. MOT17 provides diverse real-world scenes with complex motion and illumination changes, while DAVIS offers high-quality videos with rich textures and fine details, which are well suited for assessing visual fidelity and temporal consistency. For synthetic SEUD data, we apply the weather synthesis pipeline in Sec.~\ref{sec:video_weather} to clean sequences to generate time-varying rain, haze, and snow. We consider five settings to model realistic temporal evolution: (1) single-type per video with smoothly varying severity; (2) single-type with intra-video type transitions and smooth severity changes (e.g., haze$\rightarrow$rain$\rightarrow$snow); (3) fixed compound types per video with time-varying overall severity; (4) compound types whose component severities evolve independently; and (5) an open-world mixture that includes no-degradation, single-type, and compound segments with possible type-set changes. In all settings, intensity trajectories follow step, trapezoid, Gaussian, and cosine profiles that jointly modulate atmospheric transmittance and per-frame particle density.


We compare ORCANet with representative all-in-one image restoration methods, including PromptIR \cite{PIPli2023}, AirNet \cite{air2022Li}, TransWeather \cite{transweather2022Jose}, and T3DiffWeather \cite{T3diff2024chen}, as well as video restoration models such as BasicVSR++ \cite{basicvsrplus2021kelvin}, RVRT \cite{RVRT2022Liang}, and AverNet \cite{aver2024zhao}. We train all baselines using their official implementations and the best hyperparameters reported in the original papers. We use identical training and testing splits and apply the same degradation settings for fair comparison. 

For quantitative evaluation, we employ Peak Signal-to-Noise Ratio (PSNR) \cite{criterionPSNR} and Structural Similarity Index Measure (SSIM) \cite{criterionSSIM} to assess restoration quality. 

\subsection{Evaluation on Dynamic and Compound Degradation}

\begin{figure*}
    \centering
    \includegraphics[width=1\linewidth]{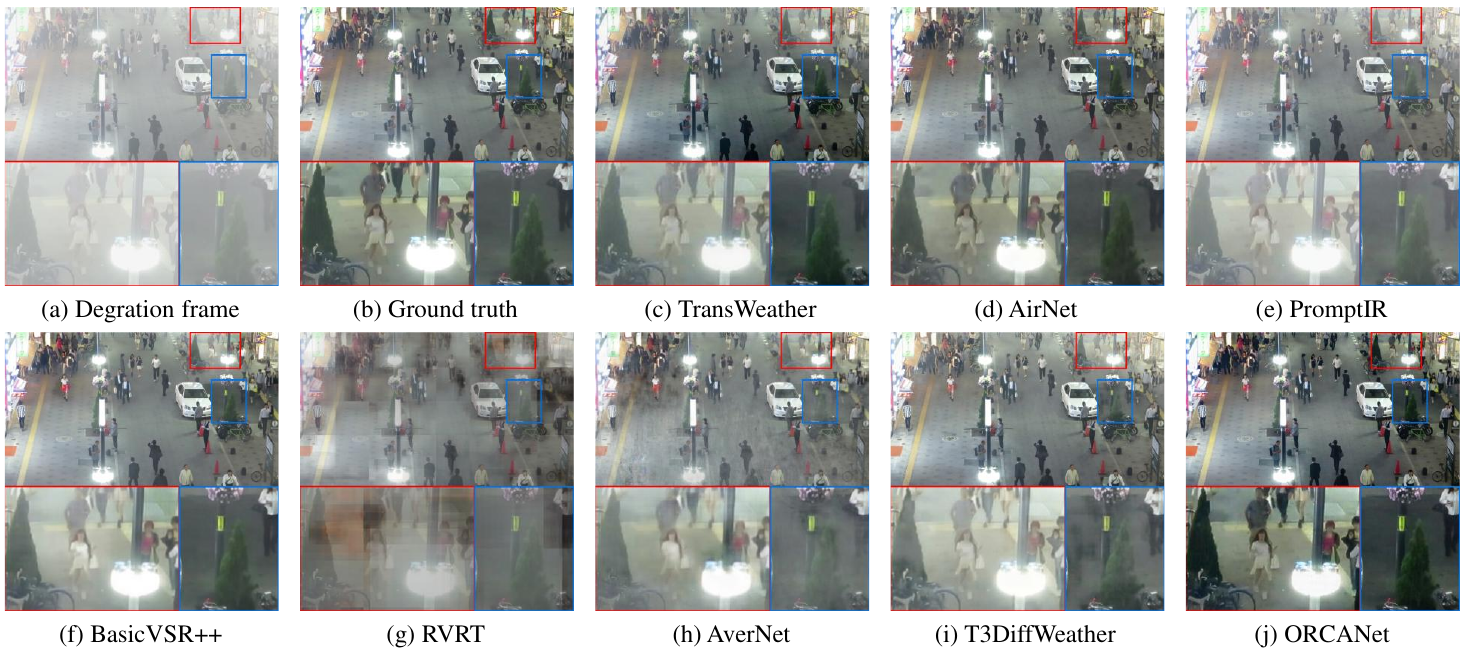}
    \caption{Comparison of visual restoration results of foggy degraded video frame.}
    \label{fig:setting2}
\end{figure*}

\begin{table*}[htbp]
\centering
\fontsize{9}{12}\selectfont
\caption{Comparison on Mot17 and DAVIS-test under three degradation settings.}
\begin{tabular}{c|cc|cc|cc|cc|cc|cc}
\toprule
\multirow{3}{*}{Method} &
\multicolumn{6}{c|}{Mot17} &
\multicolumn{6}{c}{DAVIS-test} \\
\cline{2-13}
& \multicolumn{2}{c|}{Setting 3} & \multicolumn{2}{c|}{Setting 4} & \multicolumn{2}{c|}{Setting 5}
& \multicolumn{2}{c|}{Setting 3} & \multicolumn{2}{c|}{Setting 4} & \multicolumn{2}{c}{Setting 5} \\
\cline{2-13}
& PSNR & SSIM & PSNR & SSIM & PSNR & SSIM
& PSNR & SSIM & PSNR & SSIM & PSNR & SSIM \\
\midrule
TransWeather \cite{transweather2022Jose} & 19.40 & 0.5429 & 20.94 & 0.6685 & 19.81 & 0.6050 & 20.61 & 0.5834 & 20.60 & 0.5938 & 20.90 & 0.6156 \\
AirNet \cite{air2022Li}   & 25.38 & 0.9290 & 24.66 & 0.9123 & 22.15 & 0.8761 & 21.63 & 0.8496 & 19.68 & 0.8072 & 22.91 & 0.8663 \\
PromptIR \cite{PIPli2023} & 25.93 & 0.9383 & 24.09 & 0.9139 & 28.79 & 0.9256 & 23.94 & 0.8849 & 22.45 & 0.8598 & 27.86 & 0.8960 \\
BasicVSR++ \cite{basicvsrplus2021kelvin} & 29.76 & 0.9699 & 28.09 & 0.9447 & 29.64 & 0.9636 & 28.27 & 0.9567 & 26.51 & 0.9418 & 30.06 & 0.9615 \\
RVRT \cite{RVRT2022Liang} & 18.99 & 0.8376 & 19.01 & 0.8537 & 23.36 & 0.9040 & 21.70 & 0.8536 & 19.76 & 0.8145 & 24.96 & 0.8615 \\
AverNet \cite{aver2024zhao} & 27.60 & 0.8658 & 24.40 & 0.8870 & 27.69  & 0.8816 
& 25.16 & 0.8507 & 24.96 & 0.8427 & 24.88 & 0.8464 \\
T3DiffWeather \cite{T3diff2024chen} & 30.00 & 0.9023 & 28.14 & 0.9591 & 33.65 & 0.9515
& 27.10 & 0.8876 & 25.00 & 0.8702 & 28.94 & 0.8971 \\
OCRANet (Ours)  & \textbf{32.40} & \textbf{0.9721} & \textbf{30.02} & \textbf{0.9743} & \textbf{36.06} & \textbf{0.9847}  
                & \textbf{29.24} & \textbf{0.9613} & \textbf{27.14} & \textbf{0.9450} &  \textbf{31.96} & \textbf{0.9776} \\
\bottomrule
\end{tabular}
\label{tab:compounddynamic}
\end{table*}

We first show the proposed ORCANet on dynamic single degradation scenarios, where each frame contains only one degradation type but with smoothly varying intensity in video. Setting~1 represents single-type degradations with fixed type and varying intensity across the video, while Setting~2 introduces intra-video transitions between different single-type degradations. Quantitative results on MOT17 and DAVIS-test datasets are reported in Table~\ref{tab:singledynamic}. ORCANet achieves the best performance across all settings, outperforming both all-in-one image restoration and video restoration methods by a clear margin. Specifically, our method reaches 37.88\,dB/0.9807 and 35.22\,dB/0.9795 PSNR/SSIM on MOT17 under Setting~1 and Setting~2, respectively, which surpasses the second-best method by 1.62\,dB and 2.75\,dB. On DAVIS-test, ORCANet also maintains the highest PSNR and SSIM, demonstrating consistent restoration quality under varying intensity and type transitions. We also report runtime and parameter statistics in Table~\ref{tab:singledynamic}. We measure runtime as the average per-folder processing time on the DAVIS-test set.

Figure~\ref{fig:setting1} presents qualitative results of rain degradation on the MOT17 test set. All compared methods reduce rain streaks to some extent, yet visible residues remain in most restored frames. BasicVSR++ removes fine rain streaks more effectively, while PromptIR reduces the overall contrast of rain artifacts but oversmooths details. T3DiffWeather restores near-field degradations well but leaves strong rain traces in the sky regions. In contrast, ORCANet produces the most visually clean results, where both near and distant rain streaks are nearly eliminated, and spatial details are well preserved.

Figure~\ref{fig:setting2} shows qualitative comparisons on haze-degraded videos under Setting~2. RVRT and AverNet exhibit obvious artifacts, including blocky regions and dark smearing in heavily hazed areas. AirNet and BasicVSR++ perform better in haze removal. Both methods effectively clear haze in nearby regions and maintain brightness and contrast close to the ground truth. However, AirNet fails to completely remove distant thick haze and shows noticeable brightness inconsistency across frames. BasicVSR++ preserves overall structure but loses fine texture details and leaves residual thin haze around image boundaries. In comparison, ORCANet produces the most uniform and visually consistent results. It maintains stable brightness across frames, restores fine textures, and achieves balanced haze removal in both dense and light regions. The restored frames from ORCANet also appear with slightly higher contrast than the ground truth. This can be attributed to the bright white light in the original clean street scenes, which behaves similarly to haze. Consequently, ORCANet suppresses the light scattering effect, yielding darker tones and more vivid colors that enhance perceptual quality even beyond the reference images.

\begin{figure*}
    \centering
    \includegraphics[width=1\linewidth]{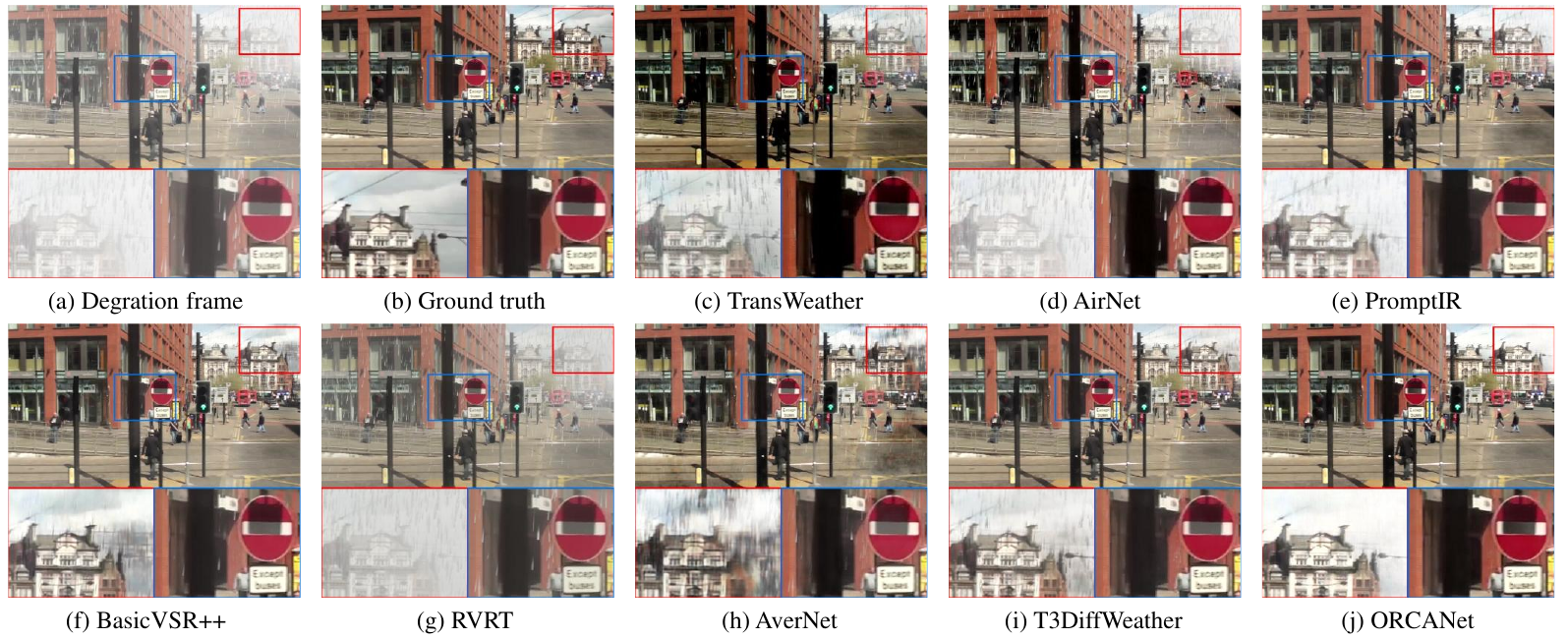}
    \caption{Comparison of visual restoration results under compound degraded video frame.}
    \label{fig:setting4}
\end{figure*}

\begin{figure*}
    \centering
    \includegraphics[width=1\linewidth]{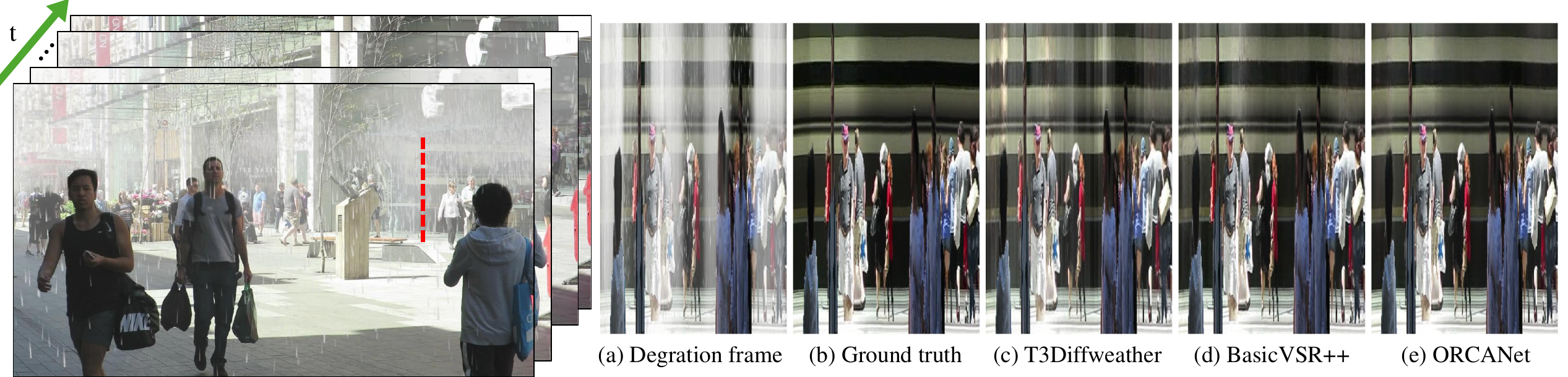}
    \caption{Temporal-axis visualization of restoration consistency. Left image illustrates the temporal variation of a sampled pixel line. (a) shows the input degraded frames. (b) shows the clean frames. (c)--(e) present the restored pixel-line slices from different methods. ORCANet achieves the most stable and coherent restoration along the temporal axis.}
    \label{fig:tempconsis}
\end{figure*}
We further evaluate ORCANet under complex mixed degradation scenarios that involve multiple and evolving degradation types. Setting~3 includes fixed compound degradations within each video, where the overall intensity varies smoothly over time. Setting~4 introduces compound degradations with independently evolving component intensities, reflecting more realistic weather dynamics. Setting~5 represents an open-world case that may contain segments of no degradation, single-type, and compound degradations, with both type composition and intensity varying freely. These three settings together form a comprehensive benchmark for evaluating model robustness under compound and time-varying conditions.

Quantitative results on MOT17 and DAVIS-test are summarized in Table~\ref{tab:compounddynamic}. ORCANet consistently achieves the best performance across all settings and datasets. On MOT17, it obtains 32.40\,dB/0.9421, 30.02\,dB/0.9743, and 36.06\,dB/0.9847 PSNR/SSIM in Settings~3–5, outperforming the second-best method T3DiffWeather by a large margin. Similar trends appear on DAVIS-test, where ORCANet maintains stable and high-quality restoration even when degradation types and severities change drastically over time. These results demonstrate the strong adaptability of ORCANet to complex and continuously evolving degradation compositions.

Figure~\ref{fig:setting4} shows qualitative results on videos with compound degradations in Setting~4. T3DiffWeather performs worse than in single-degradation cases and leaves noticeable artifacts in overlapping degradation regions. PromptIR and BasicVSR++ achieve competitive results overall. However, PromptIR retains many dark rain traces in distant sky regions and thin haze in several local areas. BasicVSR++ also leaves visible rain streaks, especially in near-field regions, while partially blurring object boundaries. Although it preserves sky details such as power lines and clouds, texture clarity decreases and dark smearing appears. In comparison, ORCANet delivers the best perceptual quality. While a few fine details in the clouds are softened, the overall visual appearance is cleaner, the degradation is well removed, and the restored scene exhibits natural color balance and smooth transitions across the frame. These results indicate that ORCANet maintains robustness and stability under complex, time-varying compound degradations.

\begin{figure*}
    \centering
    \includegraphics[width=1\linewidth]{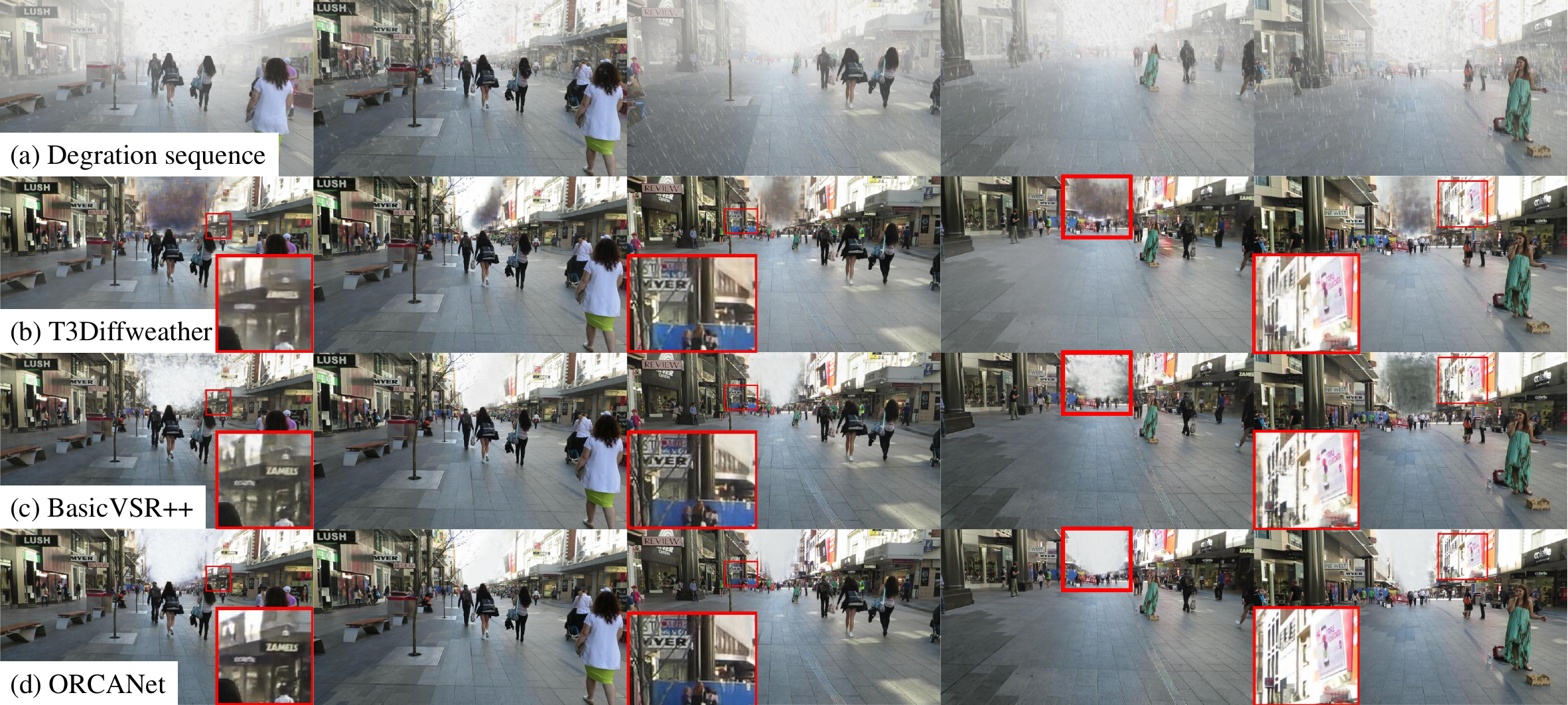}
    \caption{Visual comparison under Setting~4 with time-varying type and intensity. ORCANet maintains consistent restoration quality across time, while other methods exhibit instability or artifacts when degradation composition and strength change.}
    \label{fig:flowfigs}
\end{figure*}

Overall, the results across Settings~1–5 on MOT17 and DAVIS-test show that ORCANet handles both dynamic single-type and complex mixed degradations effectively. It sustains high PSNR/SSIM, stable brightness over time, and uniform removal in both dense and light regions. Compared with PromptIR, AirNet, BasicVSR++, RVRT, T3DiffWeather, and AverNet, ORCANet reduces artifacts such as blockiness, dark smearing, residual streaks, and thin haze, while better preserving textures and natural color balance. These quantitative and qualitative findings confirm superior restoration quality, temporal consistency, and visual fidelity under evolving intensities, intra-video type transitions, and compound compositions.

\subsection{Degradation Frame Stability and Consistency}

Temporal stability and inter-frame consistency are essential for achieving perceptually coherent video restoration. To evaluate these aspects, we analyze how the restored results behave along the temporal dimension under varying degradation strengths and types.

\begin{figure}
    \centering
    \includegraphics[width=1\linewidth]{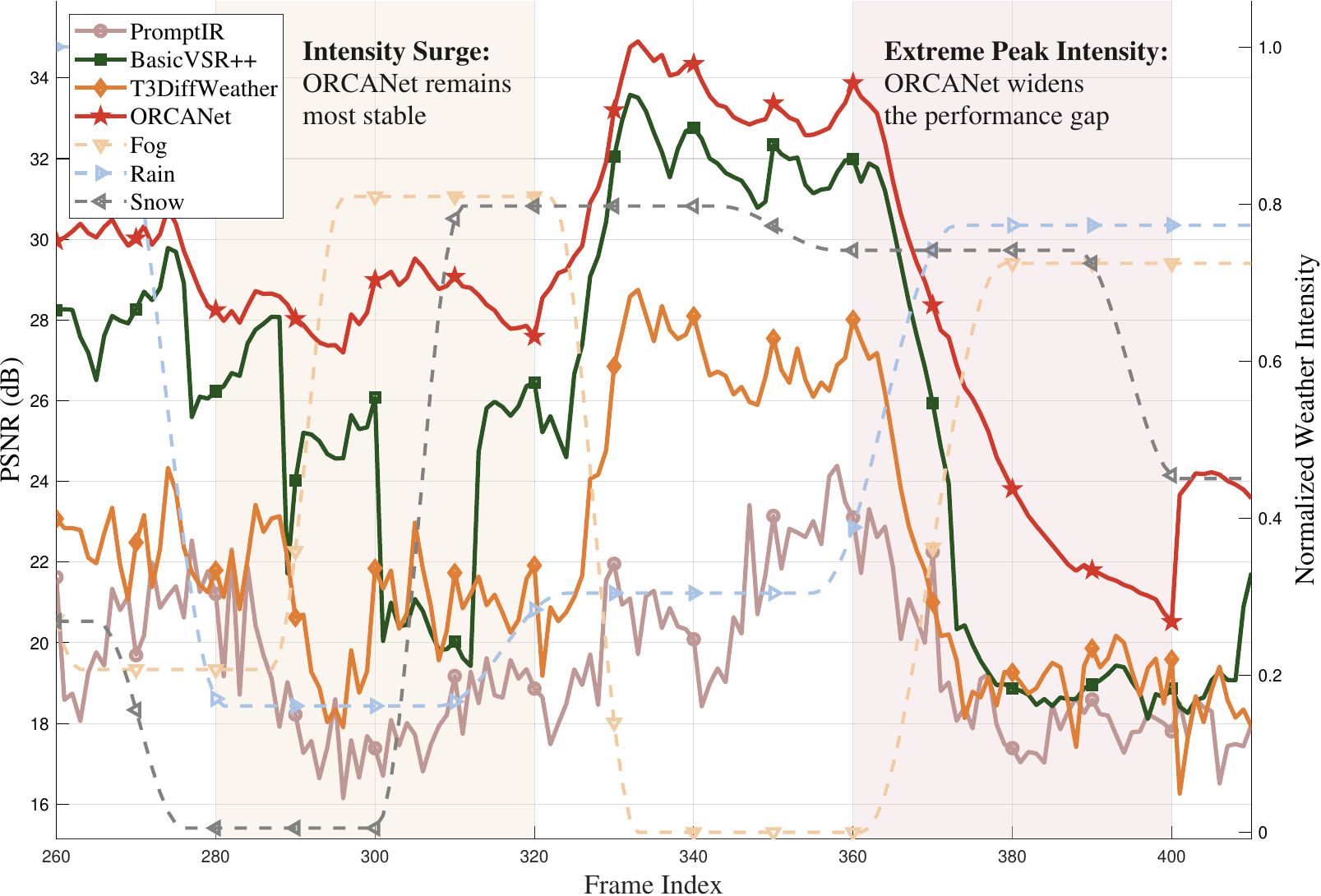}
    \caption{Frame-wise PSNR and normalized weather intensity on the setting4 MOT17 test sequence. ORCANet maintains the most stable performance under varying degradation levels.}
    \label{fig:framemetric}
\end{figure}

Figure~\ref{fig:tempconsis} presents a temporal-axis visualization. We extract a vertical pixel line from the video and display its temporal evolution. Figure~\ref{fig:tempconsis}(a) shows the degradation variation of this pixel line over time, while Figures~\ref{fig:tempconsis}(c)--(e) illustrate the restored results from different methods. Both BasicVSR++ and T3DiffWeather fail to completely remove the compound degradations in several frames, producing visible intensity fluctuations and residual streaks. In contrast, ORCANet maintains stable pixel-level consistency and yields the most temporally coherent restoration throughout the sequence.

Figure~\ref{fig:flowfigs} shows visual results under Setting~4, where both degradation type and intensity change continuously. We compare the visual results of T3DiffWeather, BasicVSR++, and ORCANet at different moments. T3DiffWeather exhibits strong sensitivity to degradation intensity; its performance drops sharply when compound degradations become severe, and texture sharpness varies across frames. T3DiffWeather processes each frame independently, which makes it highly sensitive to temporal variation in both type and intensity, resulting in inconsistent removal and occasional dark artifacts in distant regions. BasicVSR++ performs better in some segments but produces severe dark smearing in dense haze regions due to its 7-frame inference window. It also fails to remove near-field rain streaks completely.  ORCANet achieves the most uniform visual quality across time. It handles intensity transitions and type changes smoothly, maintaining stable texture details and consistent brightness over the entire sequence.

\subsection{Real-World Video Restoration Results}
\begin{figure*}
    \centering
    \includegraphics[width=1\linewidth]{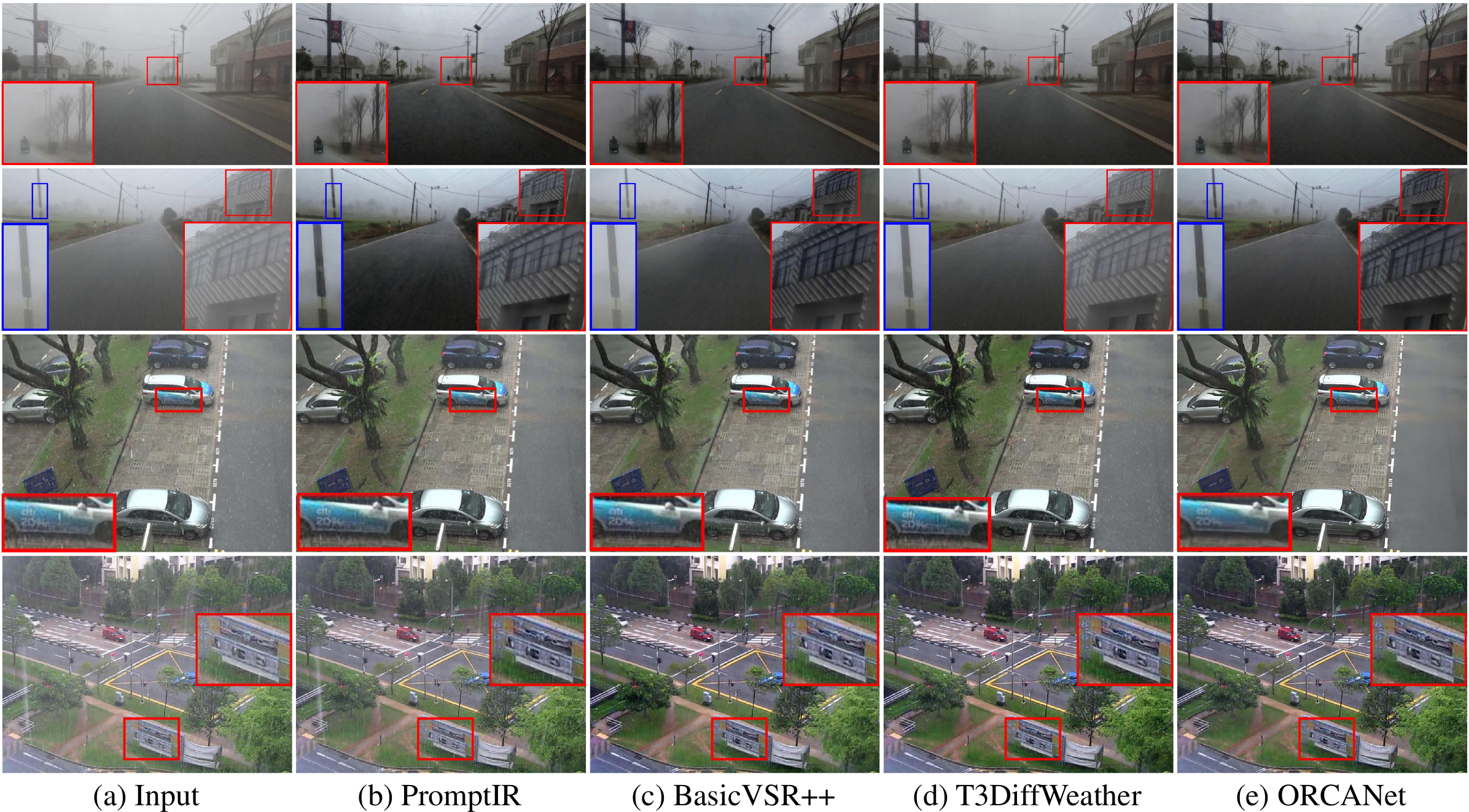}
    \caption{Visual comparisons on real-world video restoration results from GoProHazy~\cite{goprohaze2025fan} and NTURain~\cite{NTURain2018Chen}. Columns (a)--(e) show the input and the results of PromptIR, BasicVSR++, T3DiffWeather, and ORCANet. Please zoom in for clearer view.}
    \label{fig:real}
\end{figure*}
We evaluate real-world restoration on videos from GoProHazy~\cite{goprohaze2025fan} and NTURain~\cite{NTURain2018Chen}. We compare the visual results with PromptIR, BasicVSR++, and T3DiffWeather. For PromptIR and T3DiffWeather, we test two settings. One setting uses the official pretrained weights. The other setting uses weights that we train on SEUD. We report the better result for each method. Fig.~\ref{fig:real} shows the visual comparisons. On GoProHazy, BasicVSR++ and T3DiffWeather leave visible haze in some local regions. PromptIR removes haze well, but it introduces noticeable noise in several areas. ORCANet restores the nearby regions more uniformly and it preserves local details across the frame. On NTURain, the videos contain many thin rain streaks, and some scenes also include haze. T3DiffWeather recovers richer textures, but it keeps most rain traces. BasicVSR++ removes many rain streaks, but it leaves residual streaks around complex regions, and haze removal is still uneven. ORCANet handles these compound degradations better and it produces the cleanest and most stable appearance in our comparisons.

\subsection{Ablation Studies}
\begin{table*}[t]
  \centering
  \caption{Ablation on core components of ORCANet on MOT17 and DAVIS-test under SEUD Setting~4. }
  \label{tab:ablation_core}
  \setlength{\tabcolsep}{2.8mm}
  \renewcommand{\arraystretch}{1.15}
  \begin{tabular}{lcccccccc}
    \toprule
    \multirow{2}{*}{Variant} & \multirow{2}{*}{CIED} & \multirow{2}{*}{Dynamic Prompt} & \multirow{2}{*}{Static Prompt} & \multirow{2}{*}{Bi-dir} & 
    \multicolumn{2}{c}{MOT17} & \multicolumn{2}{c}{DAVIS-test} \\
    \cmidrule(lr){6-7} \cmidrule(lr){8-9}
     & & & & & PSNR & SSIM & PSNR & SSIM \\
    \midrule
    Baseline-A: NAFNet &  &  &  &  & 26.6910 & 0.9440 & 25.1004 & 0.8800 \\
    Baseline-B: NAFNet + Flow &  &  &  & \checkmark & 27.0109 & 0.9490 & 25.9560 & 0.9008 \\
    Baseline-C: B + CIED        & \checkmark &  &  & \checkmark & 28.9872 & 0.9626 & 26.8912 & 0.9327 \\
    Baseline-D: B + Dynamic prompt &  & \checkmark &  & \checkmark & 27.1578 & 0.9451 & 26.0794 & 0.9079 \\
    Baseline-E: B + FPG &  & \checkmark & \checkmark & \checkmark & 28.1292 & 0.9533 & 26.4934 & 0.9127 \\
    Full ORCANet              & \checkmark & \checkmark & \checkmark & \checkmark & \textbf{30.0201} & \textbf{0.9743} & \textbf{27.1443} & \textbf{0.9450} \\
    \bottomrule
  \end{tabular}
\end{table*}

\begin{figure*}
    \centering
    \includegraphics[width=1\linewidth]{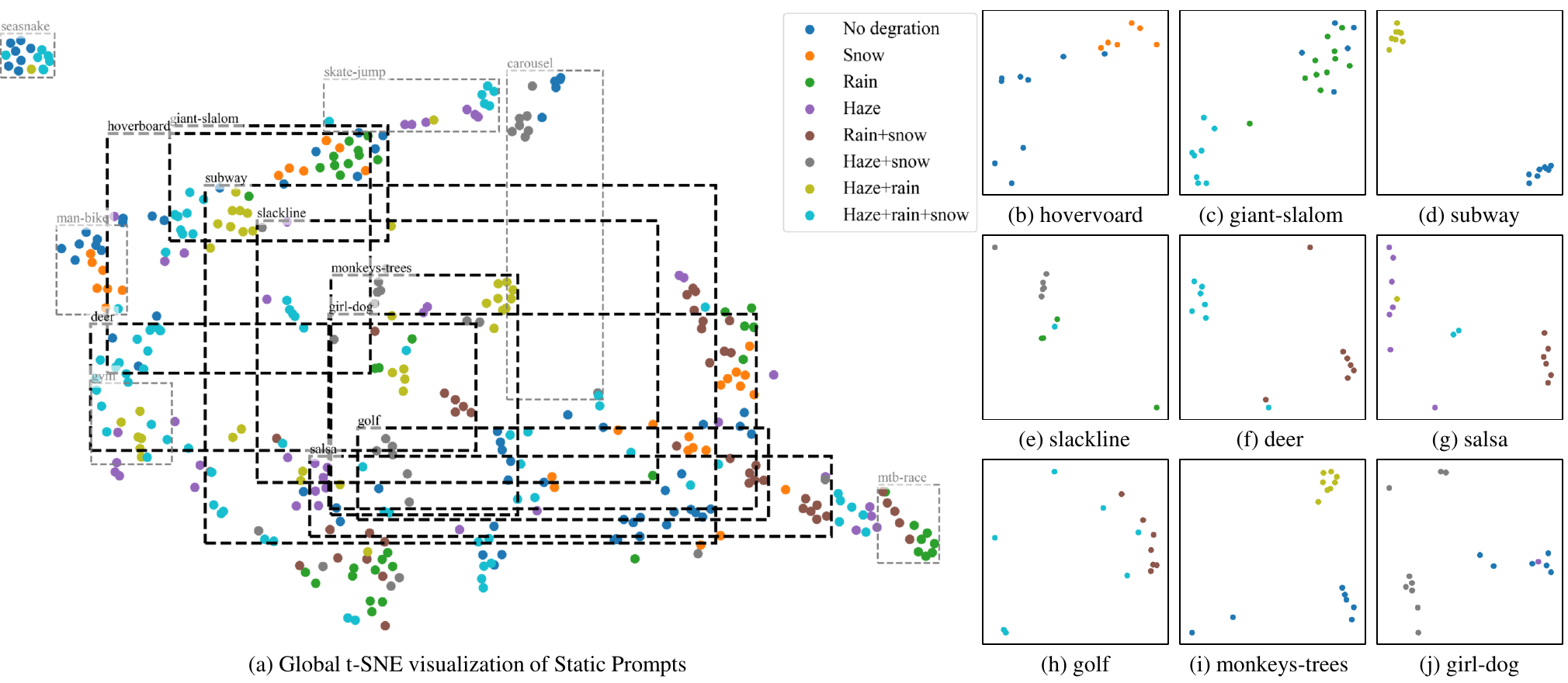}
    \caption{(a) t-SNE visualization of all static prompts in 30 DAVIS videos on Setting~5. Light dashed boxes group prompts from single videos, while dark dashed boxes highlight overlapping regions with zoomed-in views (b)-(j) on the right.}
    \label{fig:prompttsne}
\end{figure*}

Figure~\ref{fig:framemetric} shows frame-wise PSNR together with the normalized weather intensities on a MOT17 test sequence under Setting~4. From frame 280 to 320, the overall degradation strength increases. In this interval, ORCANet not only achieves the highest PSNR but also maintains the most stable curve. BasicVSR++ shows a step-like drop. It starts with performance close to ORCANet, then falls below T3DiffWeather as the degradation becomes stronger. PromptIR and T3DiffWeather both suffer a performance decline in this region and exhibit large fluctuations between adjacent frames. From frame 320 to 360, the degradation strength remains relatively stable, and all methods mainly fluctuate with scene complexity. From frame 360 to 400, the scene becomes more complex and the combined degradation reaches its peak. All methods experience a sharp performance drop. ORCANet widens its performance gap in this interval, although its PSNR reaches a local minimum around frame 400. This behavior is consistent with our inference strategy in Sec.~\ref{sec:trainstra}, where we process non-overlapping clips of 80 frames, which induces local performance dips near the clip boundaries (around frames 320 and 400). Overall, ORCANet is the most stable model under rapid changes in degradation strength.

Overall, the above analyses demonstrate that ORCANet effectively preserves temporal consistency and pixel stability under dynamic degradation evolution. It produces coherent restoration sequences that remain robust against smoothly evolving in degradation strength and type, ensuring both visual smoothness and quantitative reliability across frames.

We conduct ablation experiments to analyze the main components of ORCANet, including flow-based temporal propagation, the CIED module, and the FPG prompt module. Table~\ref{tab:ablation_core} reports results on MOT17 and DAVIS-test under SEUD Setting~4. For each variant, we remove or simplify one component while keeping the others unchanged. When we remove CIED or the static prompt branch, we also disable their corresponding loss terms. We initialize weights from the partially trained ORCANet and fine-tune them for 200{,}000 iterations on the same training data.

Baseline-A uses the single-frame NAFNet backbone without any temporal modeling. Baseline-B introduces optical-flow-based alignment and bidirectional propagation. This temporal modeling already improves PSNR on both datasets and increases SSIM on DAVIS-test, which shows that explicit alignment across frames is beneficial even without weather-specific modules.

Baseline-C further adds the CIED module on top of Baseline-B and its corresponding loss term. This variant brings the largest single gain: it improves MOT17 PSNR by almost 2\,dB and also yields clear SSIM improvements on both datasets. These results confirm that coupling coarse deweathering and intensity estimation with the backbone provides stronger and more stable features for downstream restoration.

Baseline-D uses only the dynamic prompt and does not apply any prompt-related loss, while Baseline-E enables both dynamic and static prompts with full loss supervision. Dynamic prompts alone provide a small gain over Baseline-B. Adding static prompts in Baseline-E produces a much larger improvement, which indicates that segment-level static prompts and frame-wise dynamic prompts are complementary. The full ORCANet combines CIED with FPG and the bidirectional flow-based propagation. It achieves the best performance and gains about 3.3\,dB PSNR on MOT17 and 2.2\,dB on DAVIS-test over the single-frame NAFNet baseline, which demonstrates the effectiveness of the complete design.

We further analyze the behavior of static prompts in the learned embedding space. We collect static prompts from 30 videos in Setting~5 of the DAVIS test set and project them into 2D using t-SNE. As shown in Fig.~\ref{fig:prompttsne}, each point denotes one segment-level static prompt and the color encodes the underlying degradation composition. Light dashed boxes highlight prompts from individual videos, which tend to form compact clusters. Dark dashed boxes indicate regions where prompts from multiple videos overlap; for clarity, we also show several zoomed-in views on the right. In each zoom-in, prompts from different degradations remain well separated, suggesting that static prompts capture both video-specific and degradation-aware structure in a discriminative way. 

\subsection{Limitations} \label{sec:limitai}



Although ORCANet achieves strong performance under the SEUD setting, several limitations remain. First, the current pipeline processes videos in non-overlapping segments, which leads to noticeable quality degradation near segment boundaries. This segmentation strategy simplifies memory management but introduces temporal discontinuities that affect the restoration of boundary frames. Second, the synthetic data generation assumes slow camera motion. This assumption holds for most sequences but does not cover a small portion of DAVIS videos with fast camera movement. Handling such cases may require more advanced synthesis strategies and restoration architectures that better model strong spatial--temporal coupling. In addition, the SEUD setting in this work focuses on weather degradations. The considered degradation types are limited to common cases, and the intensity functions used to simulate temporal evolution remain simplified. Future work may explore a wider range of weather processes and extend SEUD to non-weather degradation scenarios.

\section{Conclusion}
In this work, we study the problem of all-in-one video restoration under smoothly evolving and unknown degradations. We introduce the SEUD scenario, which models both temporal continuity and dynamic composition of weather degradations. To support this task, we design a flexible synthesis pipeline that generates realistic time-varying weather effects with controllable particle behavior and continuous intensity trajectories. 

We also propose ORCANet, an all-in-one restoration framework that integrates coarse intensity–aware dehazing, flow-guided bidirectional propagation, and a two-level prompt design that captures both segment-level degradation identity and frame-level variation.
Extensive experiments show that ORCANet achieves superior performance across a wide range of dynamic and compound degradation conditions. The model produces stable and temporally consistent results and demonstrates strong generalization when degradation strength and type change over time. These results confirm the importance of modeling temporal continuity and degradation-aware prompts in video all-in-one restoration. Future work will further broaden the scope of degradation types, and develop lightweight variants for real-time applications.

\bibliographystyle{ieeetr}

\begin{thebibliography}{10}

\bibitem{aiowea2020Li}
R.~Li, R.~T. Tan, and L.-F. Cheong, ``All in one bad weather removal using architectural search,'' in {\em Proc. IEEE Conf. Comput. Vis. Pattern Recognit.}, pp.~3172--3182, 2020.

\bibitem{transweather2022Jose}
J.~M. Jose~Valanarasu, R.~Yasarla, and V.~M. Patel, ``Transweather: Transformer-based restoration of images degraded by adverse weather conditions,'' in {\em Proc. IEEE Conf. Comput. Vis. Pattern Recognit.}, pp.~2343--2353, 2022.

\bibitem{air2022Li}
B.~Li, X.~Liu, P.~Hu, Z.~Wu, J.~Lv, and X.~Peng, ``All-in-one image restoration for unknown corruption,'' in {\em Proc. IEEE Conf. Comput. Vis. Pattern Recognit.}, pp.~17431--17441, 2022.

\bibitem{PIPli2023}
Z.~Li, Y.~Lei, C.~Ma, J.~Zhang, and H.~Shan, ``Prompt-in-prompt learning for universal image restoration,'' {\em arXiv preprint arXiv:2312.05038}, 2023.

\bibitem{wdiffusion2023ozan}
O.~Özdenizci and R.~Legenstein, ``Restoring vision in adverse weather conditions with patch-based denoising diffusion models,'' {\em IEEE Trans. Pattern Anal. Mach. Intell.}, vol.~45, no.~8, pp.~10346--10357, 2023.

\bibitem{PerIR25zhang}
X.~Zhang, J.~Ma, G.~Wang, Q.~Zhang, H.~Zhang, and L.~Zhang, ``Perceive-ir: Learning to perceive degradation better for all-in-one image restoration,'' {\em IEEE Trans. Image Process.}, pp.~1--1, 2025.

\bibitem{UniUIR25zhang}
X.~Zhang, H.~Zhang, G.~Wang, Q.~Zhang, L.~Zhang, and B.~Du, ``Uniuir: Considering underwater image restoration as an all-in-one learner,'' {\em IEEE Trans. Image Process.}, vol.~34, pp.~6963--6977, 2025.

\bibitem{relation25Li}
B.~Li, Y.~Gou, W.~Wang, P.~Hu, W.~Zuo, and X.~Peng, ``Relationship quantification of image degradations,'' {\em IEEE Trans. Pattern Anal. Mach. Intell.}, vol.~47, no.~8, pp.~7081--7092, 2025.

\bibitem{MFD25xiao}
Y.~Xiao, Q.~Yuan, K.~Jiang, Y.~Chen, S.~Wang, and C.-W. Lin, ``Multi-axis feature diversity enhancement for remote sensing video super-resolution,'' {\em IEEE Trans. Image Process.}, vol.~34, pp.~1766--1778, 2025.

\bibitem{Ye2022ECCVDensity}
T.~Ye, M.~Jiang, Y.~Zhang, L.~Chen, E.~Chen, P.~Chen, and Z.~Lu, ``Perceiving and modeling density is all you need for image dehazing,'' {\em arXiv preprint arXiv:2111.09733}, 2021.

\bibitem{Song2022VITDehaze}
Y.~Song, Z.~He, H.~Qian, and X.~Du, ``Vision transformers for single image dehazing,'' {\em IEEE Trans. Image Process.}, vol.~32, pp.~1927--1941, 2023.

\bibitem{D42022yang}
Y.~Yang, C.~Wang, R.~Liu, L.~Zhang, X.~Guo, and D.~Tao, ``Self-augmented unpaired image dehazing via density and depth decomposition,'' in {\em Proc. IEEE Conf. Comput. Vis. Pattern Recognit.}, pp.~2027--2036, 2022.

\bibitem{BMVC2024Kirillova}
N.~Kirillova, M.~J. Mirza, H.~Bischof, and H.~Possegger, ``Into the fog: Evaluating robustness of multiple object tracking,'' in {\em Proc. Brit. Mach. Vis. Conf.}, BMVA, 2024.

\bibitem{UCL24wang}
Y.~Wang, X.~Yan, F.~L. Wang, H.~Xie, W.~Yang, X.-P. Zhang, J.~Qin, and M.~Wei, ``Ucl-dehaze: Toward real-world image dehazing via unsupervised contrastive learning,'' {\em IEEE Trans. Image Process.}, vol.~33, pp.~1361--1374, 2024.

\bibitem{sid2021yang}
W.~Yang, R.~T. Tan, S.~Wang, Y.~Fang, and J.~Liu, ``Single image deraining: From model-based to data-driven and beyond,'' {\em IEEE Trans. Pattern Anal. Mach. Intell.}, vol.~43, no.~11, pp.~4059--4077, 2021.

\bibitem{ad2016you}
S.~You, R.~T. Tan, R.~Kawakami, Y.~Mukaigawa, and K.~Ikeuchi, ``Adherent raindrop modeling, detection and removal in video,'' {\em IEEE Trans. Pattern Anal. Mach. Intell.}, vol.~38, no.~9, pp.~1721--1733, 2016.

\bibitem{hearain2019li}
R.~Li, L.-F. Cheong, and R.~T. Tan, ``Heavy rain image restoration: Integrating physics model and conditional adversarial learning,'' in {\em Proc. IEEE Conf. Comput. Vis. Pattern Recognit.}, pp.~1633--1642, 2019.

\bibitem{desnow2018liu}
Y.-F. Liu, D.-W. Jaw, S.-C. Huang, and J.-N. Hwang, ``Desnownet: Context-aware deep network for snow removal,'' {\em IEEE Trans. Image Process.}, vol.~27, no.~6, pp.~3064--3073, 2018.

\bibitem{Qian2018Raindrop}
R.~Qian, R.~T. Tan, W.~Yang, J.~Su, and J.~Liu, ``Attentive generative adversarial network for raindrop removal from a single image,'' {\em arXiv preprint arXiv:1711.10098}, 2018.

\bibitem{Quan2021OneGo}
R.~Quan, X.~Yu, Y.~Liang, and Y.~Yang, ``Removing raindrops and rain streaks in one go,'' in {\em Proc. IEEE Conf. Comput. Vis. Pattern Recognit.}, pp.~9143--9152, 2021.

\bibitem{Chen2020JSTASR}
W.-T. Chen, H.-Y. Fang, J.-J. Ding, C.-C. Tsai, and S.-Y. Kuo, ``Jstasr: Joint size and transparency-aware snow removal algorithm based on modified partial convolution and veiling effect removal,'' in {\em Proc. Eur. Conf. Comput. Vis.}, pp.~754--770, 2020.

\bibitem{deepse2021zhang}
K.~Zhang, R.~Li, Y.~Yu, W.~Luo, and C.~Li, ``Deep dense multi-scale network for snow removal using semantic and depth priors,'' {\em IEEE Trans. Image Process.}, vol.~30, pp.~7419--7431, 2021.

\bibitem{dis2023quan}
Y.~Quan, X.~Tan, Y.~Huang, Y.~Xu, and H.~Ji, ``Image desnowing via deep invertible separation,'' {\em IEEE Trans. Circuits Syst. Video Technol.}, vol.~33, no.~7, pp.~3133--3144, 2023.

\bibitem{famamba2024xiao}
Y.~Xiao, Q.~Yuan, K.~Jiang, Y.~Chen, Q.~Zhang, and C.-W. Lin, ``Frequency-assisted mamba for remote sensing image super-resolution,'' {\em arXiv preprint arXiv:2405.04964}, 2024.

\bibitem{gou2022multi}
Y.~Gou, P.~Hu, J.~Lv, J.~T. Zhou, and X.~Peng, ``Multi-scale adaptive network for single image denoising,'' {\em Proc. Adv. Neural Inform. Process. Syst.}, vol.~35, pp.~14099--14112, 2022.

\bibitem{TTST24xiao}
Y.~Xiao, Q.~Yuan, K.~Jiang, J.~He, C.-W. Lin, and L.~Zhang, ``Ttst: A top-k token selective transformer for remote sensing image super-resolution,'' {\em IEEE Trans. Image Process.}, vol.~33, pp.~738--752, 2024.

\bibitem{gou2023rethinking}
Y.~Gou, P.~Hu, J.~Lv, H.~Zhu, and X.~Peng, ``Rethinking image super resolution from long-tailed distribution learning perspective,'' in {\em Proc. IEEE Conf. Comput. Vis. Pattern Recognit.}, pp.~14327--14336, 2023.

\bibitem{AFD23zhang}
X.~Zhang, N.~Cai, H.~Zhang, Y.~Zhang, J.~Di, and W.~Lin, ``Afd-former: A hybrid transformer with asymmetric flow division for synthesized view quality enhancement,'' {\em IEEE Trans. Circuits Syst. Video Technol.}, vol.~33, no.~8, pp.~3786--3798, 2023.

\bibitem{basicvsrplus2021kelvin}
K.~C.~K. Chan, S.~Zhou, X.~Xu, and C.~C. Loy, ``Basicvsr++: Improving video super-resolution with enhanced propagation and alignment,'' {\em arXiv preprint arXiv:2104.13371}, 2021.

\bibitem{RVRT2022Liang}
J.~Liang, Y.~Fan, X.~Xiang, R.~Ranjan, E.~Ilg, S.~Green, J.~Cao, K.~Zhang, R.~Timofte, and L.~V. Gool, ``Recurrent video restoration transformer with guided deformable attention,'' in {\em Proc. Adv. Neural Inform. Process. Syst.}, vol.~35, pp.~378--393, 2022.

\bibitem{aver2024zhao}
H.~Zhao, L.~Tian, X.~Xiao, P.~Hu, Y.~Gou, and X.~Peng, ``Avernet: All-in-one video restoration for time-varying unknown degradations,'' in {\em Proc. Adv. Neural Inform. Process. Syst.}, vol.~37, pp.~127296--127316, 2024.

\bibitem{textp2024Marcos}
M.~V. Conde, G.~Geigle, and R.~Timofte, ``Instructir: High-quality image restoration following human instructions,'' {\em arXiv preprint arXiv:2401.16468}, 2024.

\bibitem{textp2024Hao}
H.~Yang, L.~Pan, Y.~Yang, and W.~Liang, ``Language-driven all-in-one adverse weather removal,'' in {\em Proc. IEEE Conf. Comput. Vis. Pattern Recognit.}, pp.~24902--24912, 2024.

\bibitem{textp2025Yan}
Q.~Yan, A.~Jiang, K.~Chen, L.~Peng, Q.~Yi, and C.~Zhang, ``Textual prompt guided image restoration,'' {\em Engineering Applications of Artificial Intelligence}, vol.~155, p.~110981, 2025.

\bibitem{mulmodel2024Ai}
Y.~Ai, H.~Huang, X.~Zhou, J.~Wang, and R.~He, ``Multimodal prompt perceiver: Empower adaptiveness, generalizability and fidelity for all-in-one image restoration,'' in {\em Proc. IEEE Conf. Comput. Vis. Pattern Recognit.}, pp.~25432--25444, 2024.

\bibitem{T3diff2024chen}
S.~Chen, T.~Ye, K.~Zhang, Z.~Xing, Y.~Lin, and L.~Zhu, ``Teaching tailored to talent: Adverse weather restoration via prompt pool and depth-anything constraint,'' in {\em Proc. Eur. Conf. Comput. Vis.}, pp.~95--115, 2025.

\bibitem{PIRpotlapalli2023}
V.~Potlapalli, S.~W. Zamir, S.~Khan, and F.~S. Khan, ``Promptir: Prompting for all-in-one blind image restoration,'' {\em arXiv preprint arXiv:2306.13090}, 2023.

\bibitem{prores2023Ma}
J.~Ma, T.~Cheng, G.~Wang, Q.~Zhang, X.~Wang, and L.~Zhang, ``Prores: Exploring degradation-aware visual prompt for universal image restoration,'' {\em arXiv preprint arXiv:2306.13653}, 2023.

\bibitem{tap2025Wang}
H.~Wang, S.~Ji, S.~Wang, H.~Huang, X.~Jin, Q.~Zhang, and T.~Jin, ``Tap: Parameter-efficient task-aware prompting for adverse weather removal,'' {\em arXiv preprint arXiv:2508.07878}, 2025.

\bibitem{fogsim1959isra}
H.~Isra{\"e}l and F.~Kasten, {\em KOSCHMIEDERs Theorie der horizontalen Sichtweite}, pp.~7--10.
\newblock 1959.

\bibitem{yoly2021you}
B.~Li, Y.~Gou, S.~Gu, J.~Z. Liu, J.~T. Zhou, and X.~Peng, ``You only look yourself: Unsupervised and untrained single image dehazing neural network,'' {\em Int. J. Comput. Vis.}, vol.~129, no.~5, pp.~1754--1767, 2021.

\bibitem{TKL2022chen}
W.-T. Chen, Z.-K. Huang, C.-C. Tsai, H.-H. Yang, J.-J. Ding, and S.-Y. Kuo, ``Learning multiple adverse weather removal via two-stage knowledge learning and multi-contrastive regularization: Toward a unified model,'' in {\em Proc. IEEE Conf. Comput. Vis. Pattern Recognit.}, pp.~17632--17641, 2022.

\bibitem{AWRCP2023Ye}
T.~Ye, S.~Chen, J.~Bai, J.~Shi, C.~Xue, J.~Jiang, J.~Yin, E.~Chen, and Y.~Liu, ``Adverse weather removal with codebook priors,'' in {\em Proc. IEEE Int. Conf. Comput. Vis.}, pp.~12619--12630, 2023.

\bibitem{twostage2023zhu}
Y.~Zhu, T.~Wang, X.~Fu, X.~Yang, X.~Guo, J.~Dai, Y.~Qiao, and X.~Hu, ``Learning weather-general and weather-specific features for image restoration under multiple adverse weather conditions,'' in {\em Proc. IEEE Conf. Comput. Vis. Pattern Recognit.}, pp.~21747--21758, June 2023.

\bibitem{ViWS2023Yang}
Y.~Yang, A.~I. Aviles-Rivero, H.~Fu, Y.~Liu, W.~Wang, and L.~Zhu, ``Video adverse-weather-component suppression network via weather messenger and adversarial backpropagation,'' in {\em Proc. IEEE Int. Conf. Comput. Vis.}, pp.~13154--13164, 2023.

\bibitem{difftsc2024yang}
Y.~Yang, H.~Wu, A.~I. Aviles-Rivero, Y.~Zhang, J.~Qin, and L.~Zhu, ``Genuine knowledge from practice: Diffusion test-time adaptation for video adverse weather removal,'' in {\em Proc. IEEE Conf. Comput. Vis. Pattern Recognit.}, pp.~25606--25616, 2024.

\bibitem{visl2025xu}
J.~Xu, M.~Wu, X.~Hu, C.-W. Fu, Q.~Dou, and P.-A. Heng, ``Towards real-world adverse weather image restoration: Enhancing clearness and semantics with vision-language models,'' in {\em Proc. Eur. Conf. Comput. Vis.}, pp.~147--164, 2025.

\bibitem{cudn2023Cheng}
Y.~Cheng, M.~Shao, Y.~Wan, Y.~Qiao, W.~Zuo, and D.~Meng, ``Cross-consistent deep unfolding network for adaptive all-in-one video restoration,'' {\em arXiv preprint arXiv:2309.01627}, 2023.

\bibitem{MaIR2025Li}
B.~Li, H.~Zhao, W.~Wang, P.~Hu, Y.~Gou, and X.~Peng, ``Mair: A locality- and continuity-preserving mamba for image restoration,'' in {\em Proc. IEEE Conf. Comput. Vis. Pattern Recognit.}, pp.~7491--7501, June 2025.

\bibitem{depany2024yang}
L.~Yang, B.~Kang, Z.~Huang, X.~Xu, J.~Feng, and H.~Zhao, ``Depth anything: Unleashing the power of large-scale unlabeled data,'' in {\em Proc. IEEE Conf. Comput. Vis. Pattern Recognit.}, pp.~10371--10381, 2024.

\bibitem{Spy2017Anurag}
A.~Ranjan and M.~J. Black, ``Optical flow estimation using a spatial pyramid network,'' in {\em Proc. IEEE Conf. Comput. Vis. Pattern Recognit.}, pp.~2720--2729, 2017.

\bibitem{pixshuf2016shi}
W.~Shi, J.~Caballero, F.~Huszár, J.~Totz, A.~P. Aitken, R.~Bishop, D.~Rueckert, and Z.~Wang, ``Real-time single image and video super-resolution using an efficient sub-pixel convolutional neural network,'' in {\em Proc. IEEE Conf. Comput. Vis. Pattern Recognit.}, pp.~1874--1883, 2016.

\bibitem{simplebaselinesimagerestoration2022chen}
L.~Chen, X.~Chu, X.~Zhang, and J.~Sun, ``Simple baselines for image restoration,'' {\em arXiv preprint arXiv:2204.04676}, 2022.

\bibitem{loss1994Char}
P.~Charbonnier, L.~Blanc-Feraud, G.~Aubert, and M.~Barlaud, ``Two deterministic half-quadratic regularization algorithms for computed imaging,'' in {\em Proc. IEEE Int. Conf. Image Process.}, vol.~2, pp.~168--172 vol.2, 1994.

\bibitem{adam2017kingma}
D.~P. Kingma and J.~Ba, ``Adam: A method for stochastic optimization,'' {\em arXiv preprint arXiv:1412.6980}, 2014.

\bibitem{mot2016milan}
A.~Milan, L.~Leal-Taixe, I.~Reid, S.~Roth, and K.~Schindler, ``Mot16: A benchmark for multi-object tracking,'' {\em arXiv preprint arXiv:1603.00831}, 2016.

\bibitem{Davis2016F}
F.~Perazzi, J.~Pont-Tuset, B.~McWilliams, L.~Van~Gool, M.~Gross, and A.~Sorkine-Hornung, ``A benchmark dataset and evaluation methodology for video object segmentation,'' in {\em Proc. IEEE Conf. Comput. Vis. Pattern Recognit.}, pp.~724--732, 2016.

\bibitem{criterionPSNR}
Q.~Huynh-Thu and M.~Ghanbari, ``Scope of validity of {PSNR} in image/video quality assessment,'' {\em Electronics Letters}, vol.~44, pp.~800--801, 2008.

\bibitem{criterionSSIM}
Z.~Wang, A.~Bovik, H.~Sheikh, and E.~Simoncelli, ``Image quality assessment: from error visibility to structural similarity,'' {\em IEEE Trans. Image Process.}, vol.~13, no.~4, pp.~600--612, 2004.

\bibitem{goprohaze2025fan}
J.~Fan, J.~Weng, K.~Wang, Y.~Yang, J.~Qian, J.~Li, and J.~Yang, ``Driving-video dehazing with non-aligned regularization for safety assistance,'' {\em arXiv preprint arXiv:2405.09996}, 2025.

\bibitem{NTURain2018Chen}
J.~Chen, C.-H. Tan, J.~Hou, L.-P. Chau, and H.~Li, ``Robust video content alignment and compensation for rain removal in a cnn framework,'' in {\em Proc. IEEE Conf. Comput. Vis. Pattern Recognit.}, pp.~6286--6295, 2018.

\end{thebibliography}

\begin{IEEEbiography}[{\includegraphics[width=1in,height=1.25in,clip,keepaspectratio]{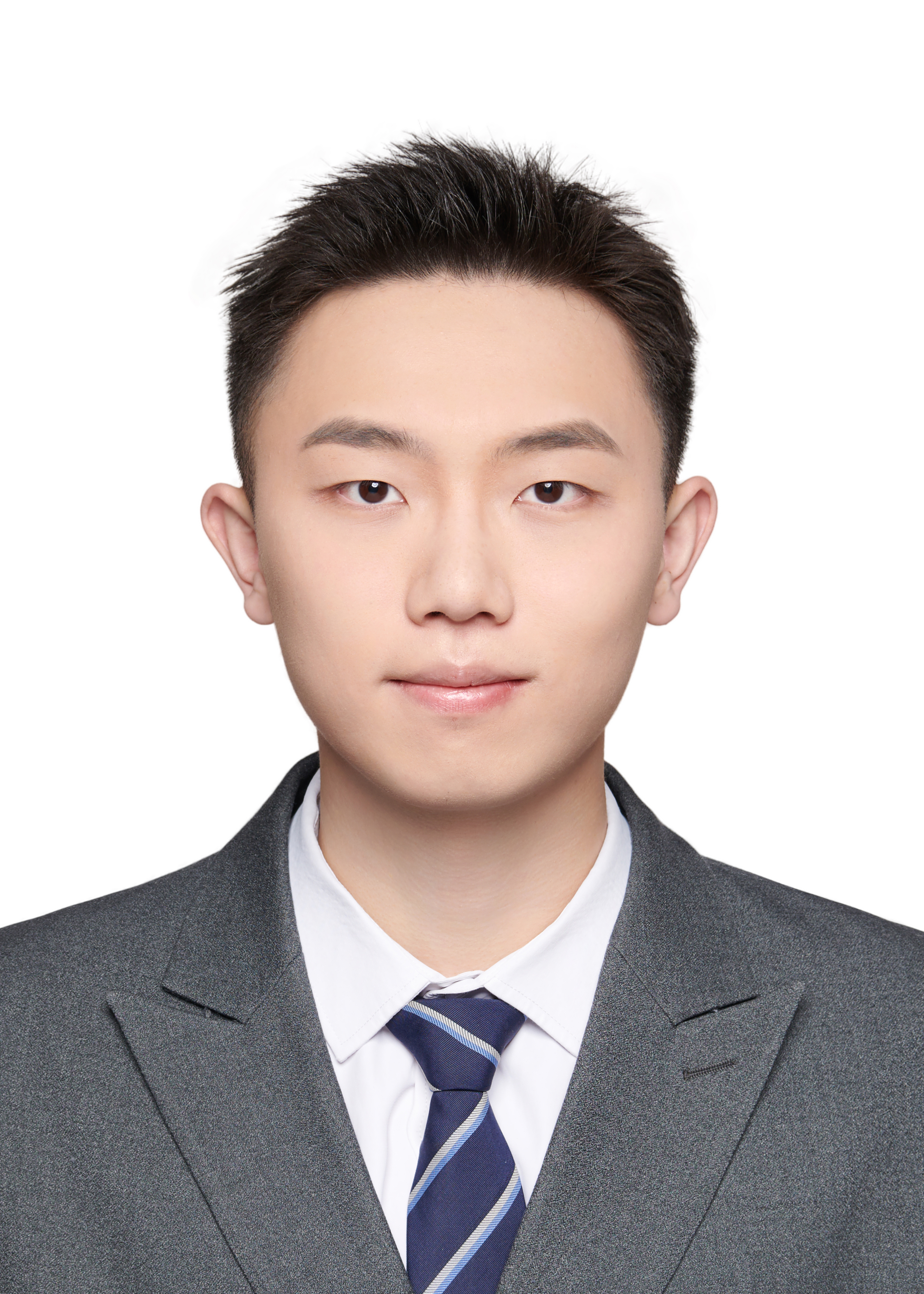}}]{Wenrui Li} (Member, IEEE) is currently an Associate Professor with the School of Computer Science, Harbin Institute of Technology. He received the B.S. degree from the School of Information and Software Engineering, University of Electronic Science and Technology of China (UESTC), Chengdu, China, in 2021, and the Ph.D. degree from the School of Computer Science, Harbin Institute of Technology (HIT), Harbin, China, in 2025. His research interests include multimedia search, low-level processing, and spiking neural networks. He has authored or co-authored more than 30 technical articles in refereed international journals and conferences.
\end{IEEEbiography}

\begin{IEEEbiography}[{\includegraphics[width=1in,height=1.25in,clip,keepaspectratio]{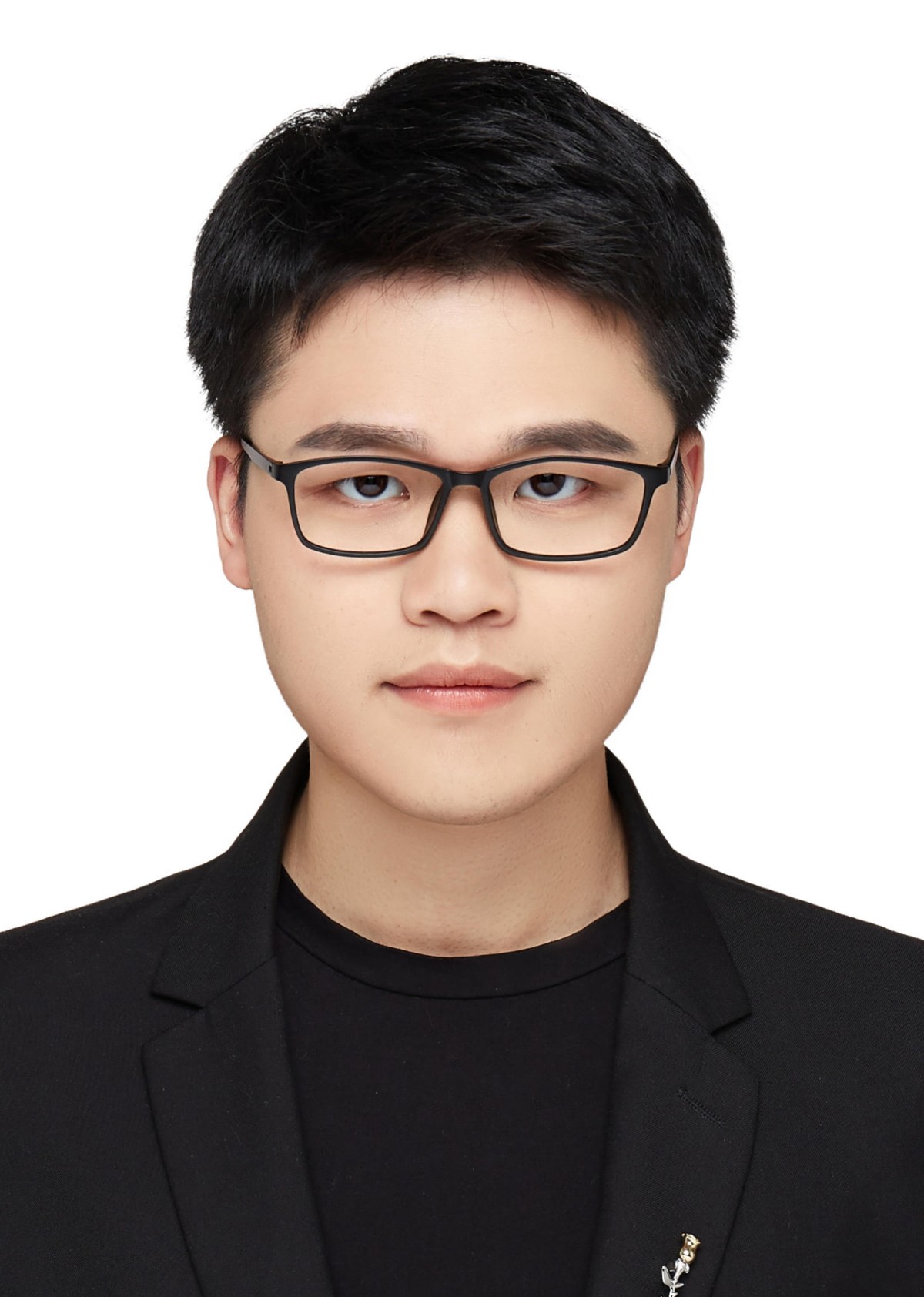}}]{Hongtao Chen} received the B.S. degree from the School of Science, Hangzhou Dianzi University (HDU), Hangzhou, China, in 2023. 
He is currently pursuing the Ph.D. degree with the School of Computer Science, Harbin Institute of Technology (HIT), Harbin, China.
His research interests include tensor completion, scientific computing, all-in-one restoration and computer vision.
\end{IEEEbiography}
\begin{IEEEbiography}[{\includegraphics[width=1in,height=1.25in,clip,keepaspectratio]{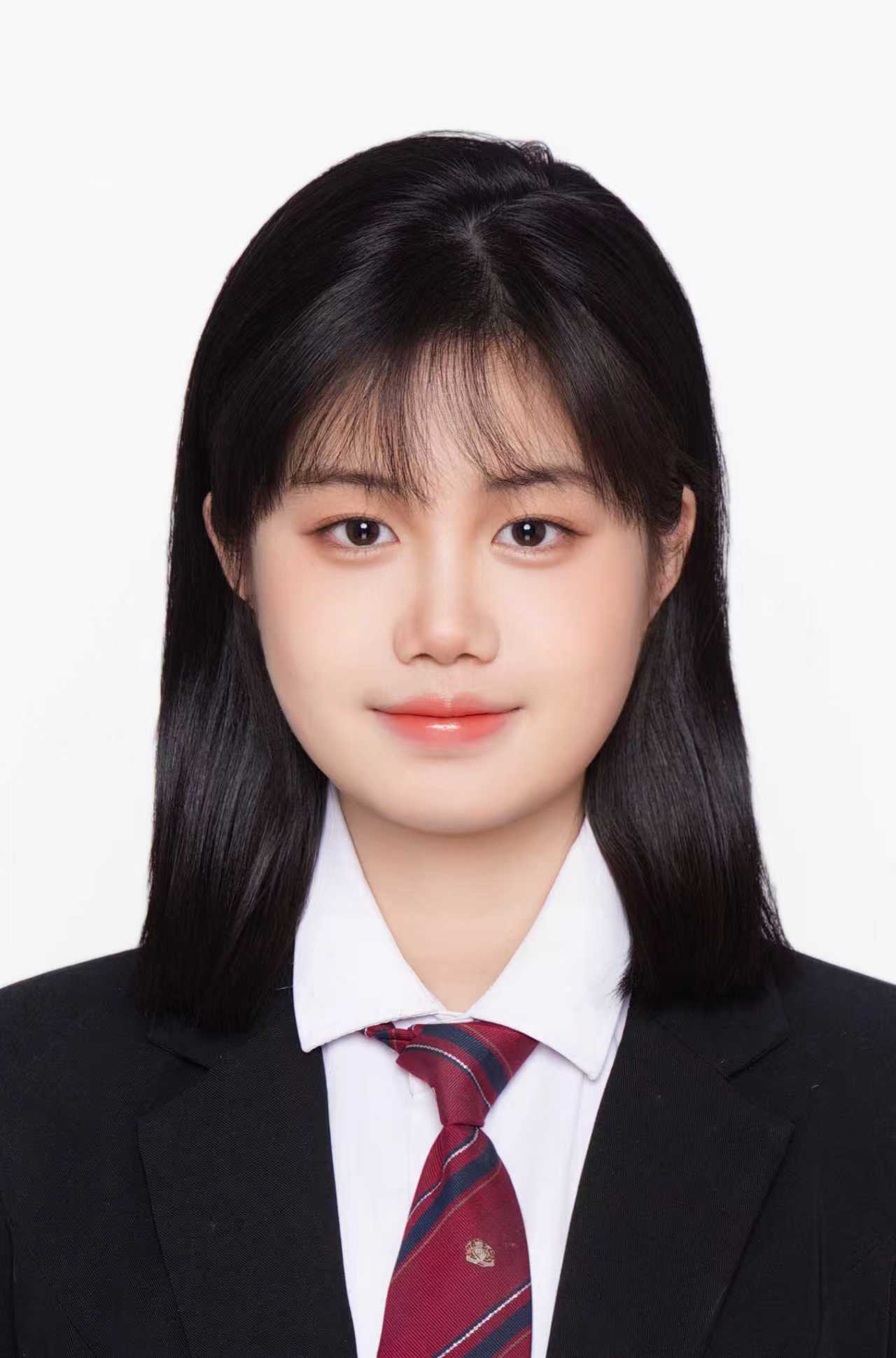}}]{Yao Xiao} received the B.S. degree from the School of Computer Science, Harbin Institute of Technology (HIT), Harbin, China, in 2024. She is currently pursuing the M.S. degree in Computer Science and Technology at HIT. Her research interests include image restoration and computer vision.
\end{IEEEbiography}
\begin{IEEEbiography}[{\includegraphics[width=1in,height=1.25in,clip,keepaspectratio]{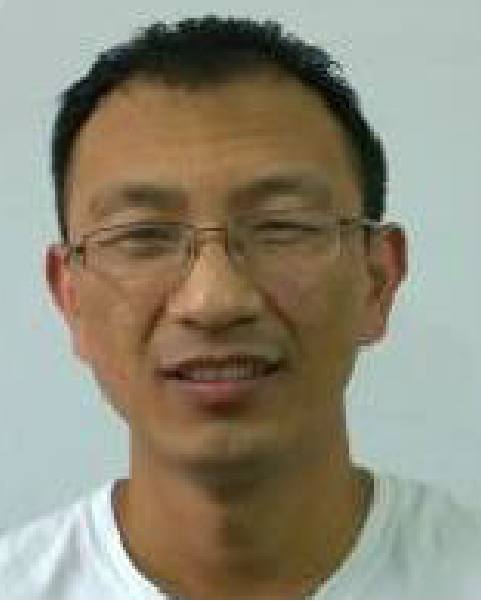}}]{Wangmeng Zuo} (Senior Member, IEEE) received the Ph.D. degree in computer application technology from Harbin Institute of Technology, Harbin, China, in 2007. He is currently a Professor with the Faculty of Computing, Harbin Institute of Technology.
He has published over 200 papers in top tier academic journals and conferences. His current research interests include low level vision, image/video generation, and multimodal understanding.
He served as an Associate Editor for IEEE TRANSACTIONS ON PATTERN ANALYSIS AND MACHINE INTELLIGENCE, IEEE TRANSACTIONS ON IMAGE PROCESSING, and SCIENCE CHINA Information Sciences.
\end{IEEEbiography}
\begin{IEEEbiography}[{\includegraphics[width=1in,height=1.25in,clip,keepaspectratio]{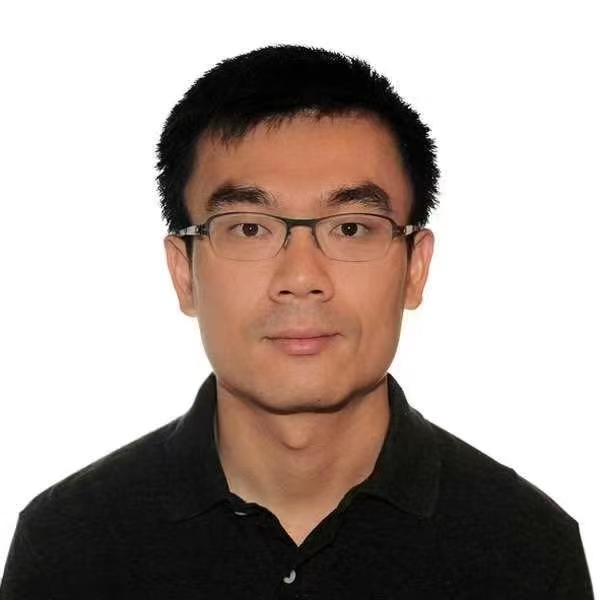}}]{Jiantao Zhou} (Senior Member, IEEE) is a Full Professor at the Department of Computer and Information Science, Faculty of Science and Technology and the State Key Laboratory of Internet of Things for Smart City, University of Macau, where he also serves as the Director for Research Services and Knowledge Transfer Office (RSKTO). He graduated from the Hong Kong University of Science and Technology in 2009 with a PhD in Electrical and Computer Engineering. He was a Fulbright Junior Scholar at the University of Illinois at Urbana-Champaign (UIUC). Professor Zhou’s research focuses on AI security, multimedia information privacy protection and forensics, and intelligent multimedia information processing. He currently serves as the Associate Editor for IEEE Trans. Multimedia (TMM) and IEEE Trans. Dependable and Secure Computing (TDSC), the top journals in the field of multimedia information processing and security, and was the Editor-in-Chief of APSIPA Newsletters. 
He is the Chair-Elect for the Multimedia Systems and Applications Technical Committee in IEEE Circuits and Systems Society, and was the TPC Co-Chair of ICME 2023 and the General-Chair of APSIPA ASC 2024. He received the 2022 Macau Science and Technology Award (Third Prize, Natural Science Award) and the 2023 Alibaba Outstanding Academic Cooperation Project Award.
\end{IEEEbiography}
\begin{IEEEbiography}[{\includegraphics[width=1in,height=1.25in,clip,keepaspectratio]{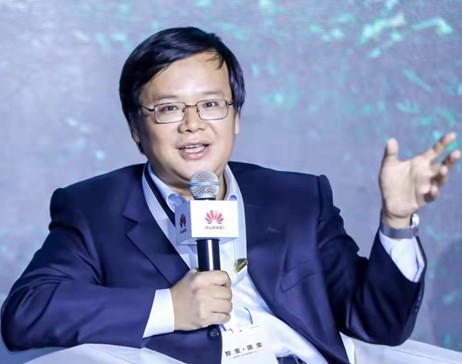}}]{Yonghong Tian} (Fellow, IEEE) is currently the Dean of the School of Electronics and Computer Engineering, a Boya Distinguished Professor with the School of
Computer Science, Peking University, China, and the Deputy Director of the Artificial Intelligence Research, Peng Cheng Laboratory, Shenzhen, China. He is the author or coauthor of over 350 technical papers in refereed journals and conferences. His
research interests include neuromorphic vision, distributed machine learning, and AI for science. He is a TPC Member of more than ten conferences, such as CVPR, ICCV, ACM KDD, AAAI, ACM MM, and ECCV. He is a Senior Member of CIE and CCF and a member of ACM. He was a recipient of the Chinese National Science Foundation for Distinguished Young Scholars in 2018, two National Science and Technology Awards, and three ministerial-level awards in China. He received the 2015 Best Paper Award for EURASIP Journal on Image and Video Processing, the Best Paper Award from IEEE BigMM 2018, and the 2022 IEEE SA Standards Medallion and SA Emerging Technology Award. He served as the TPC Co-Chair for BigMM 2015, the Technical Program Co-Chair for IEEE ICME 2015, IEEE ISM 2015, and IEEE
MIPR 2018/2019, and the General Co-Chair for IEEE MIPR 2020 and ICME 2021. He was/is an Associate Editor of IEEE TRANSACTIONS ON CIRCUITS AND SYSTEMS FOR VIDEO TECHNOLOGY from January 2018 to December 2021, IEEE TRANSACTIONS ON MULTIMEDIA from August 2014 to August
2018, IEEE Multimedia Magazine from January 2018 to August 2022, and IEEE ACCESS from January 2017 to December 2021. He co-initiated the IEEE International Conference on Multimedia Big Data.
\end{IEEEbiography}
\begin{IEEEbiography}[{\includegraphics[width=1in,height=1.25in,clip,keepaspectratio]{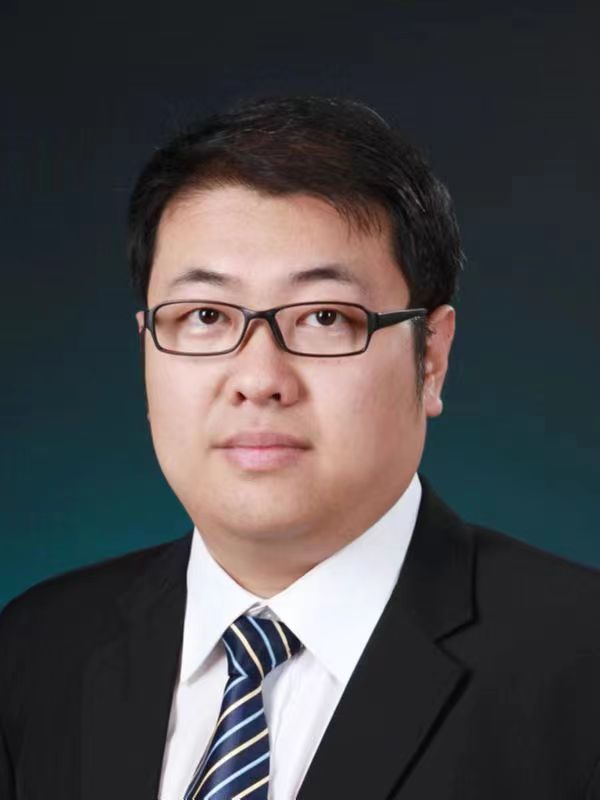}}]{Xiaopeng Fan} (Senior Member, IEEE) received the B.S. and M.S. degrees from the Harbin Institute of Technology (HIT), Harbin, China, in 2001 and 2003, respectively, and the Ph.D. degree from the Hong Kong University of Science and Technology, Hong Kong, in 2009. In 2009, he joined HIT, where he is currently a Professor. From 2003 to 2005, he was with Intel Corporation, China, as a Software Engineer. From 2011 to 2012, he was with Microsoft Research Asia, as a Visiting Researcher. From 2015 to 2016, he was with the Hong Kong University of Science and Technology, as a Research Assistant Professor. He has authored one book and more than 220 articles in refereed journals and conference proceedings. His research interests include video coding and transmission, image processing, and computer vision. He was the Program Chair of PCM2017, Chair of IEEE SGC2015, and Co-Chair of MCSN2015. He was an Associate Editor for IEEE 1857 Standard in 2012. He was the recipient of Outstanding Contributions to the Development of IEEE Standard 1857 by IEEE in 2013.
\end{IEEEbiography}
\vfill
\end{document}